\documentclass[journal=jcisd8,manuscript=article]{achemso}
\setkeys{acs}{articletitle = true}

\usepackage[version=3]{mhchem} % Formula subscripts using \ce{}
\usepackage{amsfonts}
\usepackage{makecell}
\usepackage{graphicx}
\usepackage{caption}
\usepackage{subcaption}
\usepackage{hyperref}
\usepackage{placeins}
\usepackage{longtable}
\usepackage{url}

\author{Kevin Yang}
\email{yangk@mit.edu}
\affiliation[MIT]{Computer Science and Artificial Intelligence Laboratory, MIT, Cambridge, MA 02139, United States}
\author{Kyle Swanson}
\email{swansonk@mit.edu}
\affiliation[MIT]{Computer Science and Artificial Intelligence Laboratory, MIT, Cambridge, MA 02139, United States}
\author{Wengong Jin}
\affiliation[MIT]{Computer Science and Artificial Intelligence Laboratory, MIT, Cambridge, MA 02139, United States}
\author{Connor Coley}
\affiliation[MIT2]{Department of Chemical Engineering, MIT, Cambridge, MA 02139, United States}
\author{Philipp Eiden}
\affiliation[BASF]{BASF SE, Ludwigshafen 67063, Germany}
\author{Hua Gao}
\affiliation[Amgen]{Amgen Inc., Cambridge, MA 02141, United States}
\author{Angel Guzman-Perez}
\affiliation[Amgen]{Amgen Inc., Cambridge, MA 02141, United States}
\author{Timothy Hopper}
\affiliation[Amgen]{Amgen Inc., Cambridge, MA 02141, United States}
\author{Brian Kelley}
\affiliation[Novartis]{Novartis Institutes for BioMedical Research, Cambridge, MA 02139, United States}
\author{Miriam Mathea}
\affiliation[BASF]{BASF SE, Ludwigshafen 67063, Germany}
\author{Andrew Palmer}
\affiliation[BASF]{BASF SE, Ludwigshafen 67063, Germany}
\author{Volker Settels}
\affiliation[BASF]{BASF SE, Ludwigshafen 67063, Germany}
\author{Tommi Jaakkola}
\affiliation[MIT]{Computer Science and Artificial Intelligence Laboratory, MIT, Cambridge, MA 02139, United States}
\author{Klavs Jensen}
\affiliation[MIT2]{Department of Chemical Engineering, MIT, Cambridge, MA 02139, United States}
\author{Regina Barzilay}
\affiliation[MIT]{Computer Science and Artificial Intelligence Laboratory, MIT, Cambridge, MA 02139, United States}

\title{Analyzing Learned Molecular Representations for Property Prediction}

\keywords{message passing, neural networks, machine learning, property prediction}

\begin{document}
%%%%%%%%%%%%%%%%%%%%%%%%%%%%%%%%%%%%%%%%%%%%%%%%%%%%%%%%%%%%%%%%%%%%%
%% The manuscript does not need to include \maketitle, which is
%% executed automatically.  The document should begin with an
%% abstract, if appropriate.  If one is given and should not be, the
%% contents will be gobbled.
%%%%%%%%%%%%%%%%%%%%%%%%%%%%%%%%%%%%%%%%%%%%%%%%%%%%%%%%%%%%%%%%%%%%%
\begin{abstract}

Advancements in neural machinery have led to a wide range of algorithmic solutions for molecular property prediction. Two classes of models in particular have yielded promising results: neural networks applied to computed molecular fingerprints or expert-crafted descriptors, and graph convolutional neural networks that construct a learned molecular representation by operating on the graph structure of the molecule. However, recent literature has yet to clearly determine which of these two methods is superior when generalizing to new chemical space. Furthermore, prior research has rarely examined these new models in industry research settings in comparison to existing employed models. In this paper, we benchmark models extensively on 19 public and 16 proprietary industrial datasets spanning a wide variety of chemical endpoints. In addition, we introduce a graph convolutional model that consistently matches or outperforms models using fixed molecular descriptors as well as previous graph neural architectures on both public and proprietary datasets. Our empirical findings indicate that while approaches based on these representations have yet to reach the level of experimental reproducibility, our proposed model nevertheless offers significant improvements over models currently used in industrial workflows.

\end{abstract}

\section{Introduction}

Molecular property prediction, one of the oldest cheminformatics tasks, has received new attention in light of recent advancements in deep neural networks. These architectures either operate over fixed molecular fingerprints common in traditional QSAR models, or they learn their own task-specific representations using graph convolutions \cite{duvenaud2015convolutional,Wu_2018,kearnes2016molecular,gilmer2017neural,li2015gated,kipf2016semi,defferrard2016convolutional,bruna2013spectral,coley_convolutional_2017,schutt2017quantum,battaglia2016interaction}. Both approaches are reported to yield substantial performance gains, raising state-of-the-art accuracy in property prediction.

Despite these successes, many questions remain unanswered. The first question concerns the comparison between learned molecular representations and fingerprints or descriptors. Unfortunately, current published results on this topic do not provide a clear answer. \citeauthor{Wu_2018}\cite{Wu_2018} demonstrate that convolution-based models typically outperform fingerprint-based models, while experiments reported in \citeauthor{mayr2018chembl}\cite{mayr2018chembl} report the opposite. Part of these discrepancies can be attributed to differences in evaluation setup, including the way datasets are constructed. This leads us to a broader question concerning current evaluation protocols and their capacity to measure the generalization power of a method when applied to a new chemical space, as is common in drug discovery. Unless special care is taken to replicate this distributional shift in evaluation, neural models may overfit the training data but still score highly on the test data. This is particularly true for convolutional models that can learn a poor molecular representation by memorizing the molecular scaffolds in the training data and thereby fail to generalize to new ones. Therefore, a meaningful evaluation of property prediction models needs to account explicitly for scaffold overlap between train and test data in light of generalization requirements.

In this paper, we aim to answer both of these questions by designing a comprehensive evaluation setup for assessing neural architectures. We also introduce an algorithm for property prediction that outperforms existing strong baselines across a range of datasets. The model has two distinctive features: (1) It operates over a hybrid representation that combines convolutions and descriptors. This design gives it flexibility in learning a task specific encoding, while providing a strong prior with fixed descriptors. (2) It learns to construct molecular encodings by using convolutions centered on bonds instead of atoms, thereby avoiding unnecessary loops during the message passing phase of the algorithm.

We extensively evaluate our model and other recently published neural architectures with over 850 experiments on 19 publicly available benchmarks from \citeauthor{Wu_2018}\cite{Wu_2018} and \citeauthor{mayr2018chembl}\cite{mayr2018chembl} and on 16 proprietary datasets from Amgen, Novartis, and BASF (Badische Anilin und Soda Fabrik). Our goal is to assess whether the models' performance on the public datasets and their relative ranking are representative of their ranking on the proprietary datasets. We demonstrate that under a scaffold split of training and testing data, the relative ranking of the models is consistent across the two classes of datasets. We also show that a scaffold-based split of the training and testing data is a good approximation of the temporal split commonly used in industry in terms of the relevant metrics. By contrast, a purely random split is a poor approximation to a temporal split, confirming the findings of \citeauthor{sheridan2015relative}\cite{sheridan2015relative}. To put the performance of current models in perspective, we report bounds on experimental error and show that there is still room for improving deep learning models to match the accuracy and reproducibility of screening results.

Building on the diversity of our benchmark datasets, we explore the impact of molecular representation with respect to the dataset characteristics. We find that a hybrid representation yields higher performance and generalizes better than either convolution-based or fingerprint-based models. We also note that on small datasets (up to 1000 training molecules) fingerprint models can outperform learned representations, which are negatively impacted by data sparsity. Beyond molecular representation issues, we observe that hyperparameter selection plays a crucial role in model performance, consistent with prior work\cite{shahriari2016bayes}. We show that Bayesian optimization yields a robust, automatic solution to this issue. The addition of ensembling further improves accuracy, again consistent with the literature\cite{dietterich2000ensemble}. 

Our experiments show that our model achieves consistently strong out-of-the-box performance and even stronger optimized performance across a wide variety of public and proprietary datasets. Our model achieves comparable or better performance on 11 out of 19 public datasets and on 15 out of 16 proprietary datasets compared to all baseline models. Furthermore, no single baseline model is clearly superior across the remaining 8 public datasets, and the relative performance of the baseline models often varies from dataset to dataset, whereas our model is consistently strong across datasets. These results indicate that our model, and learned molecular fingerprints in general, are applicable and ready to be used as a powerful tool for chemists actively working on drug discovery.

% Comments from CWC about the old intro that might still be applicable now:
% CWC: also commented below, but I would be careful calling fingerprints "expert-crafted"
% CWC: Lusci et al. (dx.doi.org/10.1021/ci400187y) also learn a representation from molecules as DAGs
% CWC: most people in qsar advocate for time-splits. I think it's worth mentioning that time-split cross-validations are a good way of determining generalizability but that many datasets do not have explicit information about when data were acquired. 
% CWC: Tropsha has a couple good perspectives on QSAR methods
% CWC: I think it could be worth 1-2 sentences explicitly discussing what it really means for the test set to come from the same distribution as the training set in terms of model performance & evaluation. it may not be clear to some readers why generalization would be poor when the test data differs in distribution without saying that it violates some implicit assumption 

\section{Background}
% CWC: it's not typical to have an explicit separate section for Related Work, but it should be fine for JCIM (Journal of Chemical Information and Modelling); less fine for ACS Central Science 
% New CWC comment: I think this section could be relabled "Background" and be appropriate, upon rereading it

Since the core element of our model is a graph encoder architecture, our work is closely related to previous work on graph encoders, such as those for social networks\cite{kipf2016semi,hamilton2017inductive} or for chemistry applications\cite{scarselli2009graph,bruna2013spectral,duvenaud2015convolutional,henaff2015deep,dai2016discriminative,defferrard2016convolutional,lei2017deriving,coley_convolutional_2017,kusner2017grammar,gomez2018automatic,jin2018junction,jin2018learning}.

Common approaches to molecular property prediction today involve the application of well-known models like support vector machines\cite{cortes1995support} or random forests\cite{breiman2001random} to expert-engineered descriptors or molecular fingerprints, such as the Dragon descriptors\cite{mauri2006dragon} or Morgan (ECFP) fingerprints\cite{rogers2010extended}. One direction of advancement is the use of domain expertise to improve the base feature representation of molecular descriptors\cite{swamidass2005kernels,cao2013chemopy,durant2002reoptimization,moriwaki2018mordred,mauri2006dragon} to drive better performance\cite{mayr2018chembl}. Additionally, many studies have leveraged explicit 3D atomic coordinates to improve performance further\cite{Wu_2018,schutt2017schnet,kondor2018covariant,faber2017machine,feinberg2018potentialnet}.
% CWC: what does it mean for 3D information to be nonobvious from a molecule's SMILES string? A SMILES string uniquely defines the molecule for the purposes of these models
% CWC: I would also be sure to differentiate "descriptors" like the Dragon reference and "fingerprints" like Morgan FPs, ECFPs, atom pair, torsional, etc. The descriptors tend to be the expert-engineered ones. I think most people would not describe Morgan fingerprints as "expert-engineered" 
 
The other main line of research is the optimization of the model architecture, whether the model is applied to descriptors or fingerprints\cite{mayr2018chembl,Lee201810847} or is directly applied to SMILES\cite{weininger1988smiles} strings\cite{mayr2018chembl} or the underlying graph of the molecule\cite{duvenaud2015convolutional,Wu_2018,kearnes2016molecular,gilmer2017neural,li2015gated,kipf2016semi,defferrard2016convolutional,coley_convolutional_2017,bruna2013spectral,schutt2017quantum,battaglia2016interaction}. Our model belongs to the last category of models, known as graph convolutional neural networks. In essence, such models learn their own expert feature representations directly from the data, and they have been shown to be very flexible and capable of capturing complex relationships given sufficient data\cite{gilmer2017neural,Wu_2018}. 

In a direction orthogonal to our own improvements, \citeauthor{ishiguro2019graph}\cite{ishiguro2019graph} also make a strong improvement to graph neural networks. \citeauthor{liu2018chemi}\cite{liu2018chemi} also evaluate their model against private industry datasets, but we cannot compare against their method directly owing to dataset differences\cite{liu2018chemi}. 
 
The property prediction models most similar to our own are encapsulated in the Message Passing Neural Network (MPNN) framework presented in \citeauthor{gilmer2017neural}\cite{gilmer2017neural}. We build upon this basic framework by adopting a message-passing paradigm based on updating representations of directed bonds rather than atoms. Additionally, we further improve the model by combining computed molecule-level features with the molecular representation learned by the MPNN.

\section{Methods}

We first summarize MPNNs in general using the terminology of \citeauthor{gilmer2017neural}\cite{gilmer2017neural}, and then we expand on the characteristics of Directed MPNN (D-MPNN)\cite{dai2016discriminative} used in this paper. (D-MPNN is originally called \textit{structure2vec} in \citeauthor{dai2016discriminative}\cite{dai2016discriminative}. In this paper, we refer to it as Directed MPNN to show it is a variant of the generic MPNN architecture.)

\subsection{Message Passing Neural Networks}

An MPNN is a model which operates on an undirected graph $G$ with node (atom) features $x_v$ and edge (bond) features $e_{vw}$. MPNNs operate in two phases: a \textit{message passing phase}, which transmits information across the molecule to build a neural representation of the molecule, and a \textit{readout phase}, which uses the final representation of the molecule to make predictions about the properties of interest.

More specifically, the message passing phase consists of $T$ steps. On each step $t$, hidden states $h_v^t$ and messages $m_v^t$ associated with each vertex $v$ are updated using message function $M_t$ and vertex update function $U_t$ according to:
\begin{align*}
    m_v^{t+1} &= \sum_{w \in N(v)} M_t(h_v^t, h_w^t, e_{vw}) \\
    h_v^{t+1} &= U_t(h_v^t, m_v^{t+1})
\end{align*}
where $N(v)$ is the set of neighbors of $v$ in graph $G$, and $h_v^0$ is some function of the initial atom features $x_v$.
The readout phase then uses a readout function $R$ to make a property prediction based on the final hidden states according to
\begin{equation*}
    \hat{y} = R(\{h_v^T | v \in G\}).
\end{equation*}
The output $\hat{y}$ may be either a scalar or a vector, depending on whether the MPNN is designed to predict a single property or multiple properties (in a multitask setting).

During training, the network takes molecular graphs as input and makes an output prediction for each molecule. A loss function is computed based on the predicted outputs and the ground truth values, and the gradient of the loss is backpropagated through the readout phase and the message passing phase. The entire model is trained end-to-end. 

\subsection{Directed MPNN}

\begin{figure}
    \centering
    \begin{minipage}{.45\textwidth}
        \begin{subfigure}{\textwidth}
            \includegraphics[width=\linewidth]{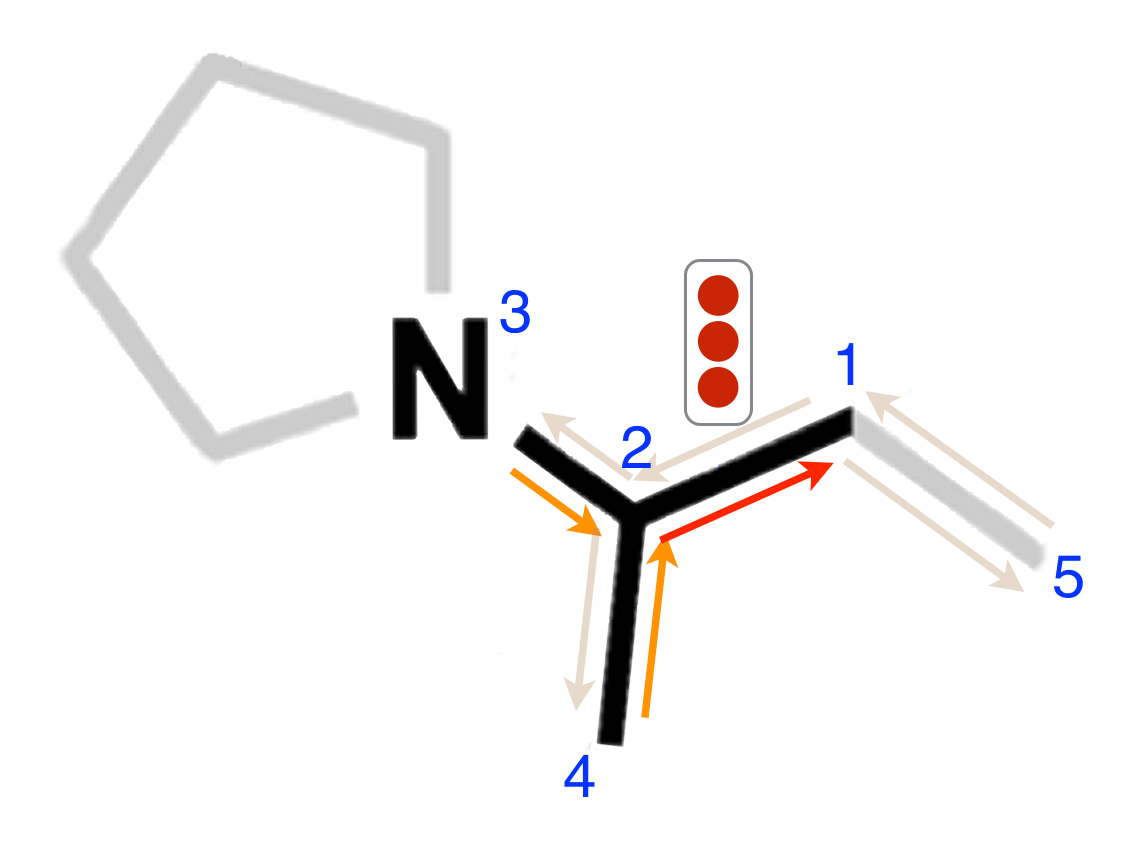}
            \caption{}
            \label{fig:bond_message_passinga}
        \end{subfigure}
        \begin{subfigure}{\textwidth}
            \includegraphics[width=\linewidth]{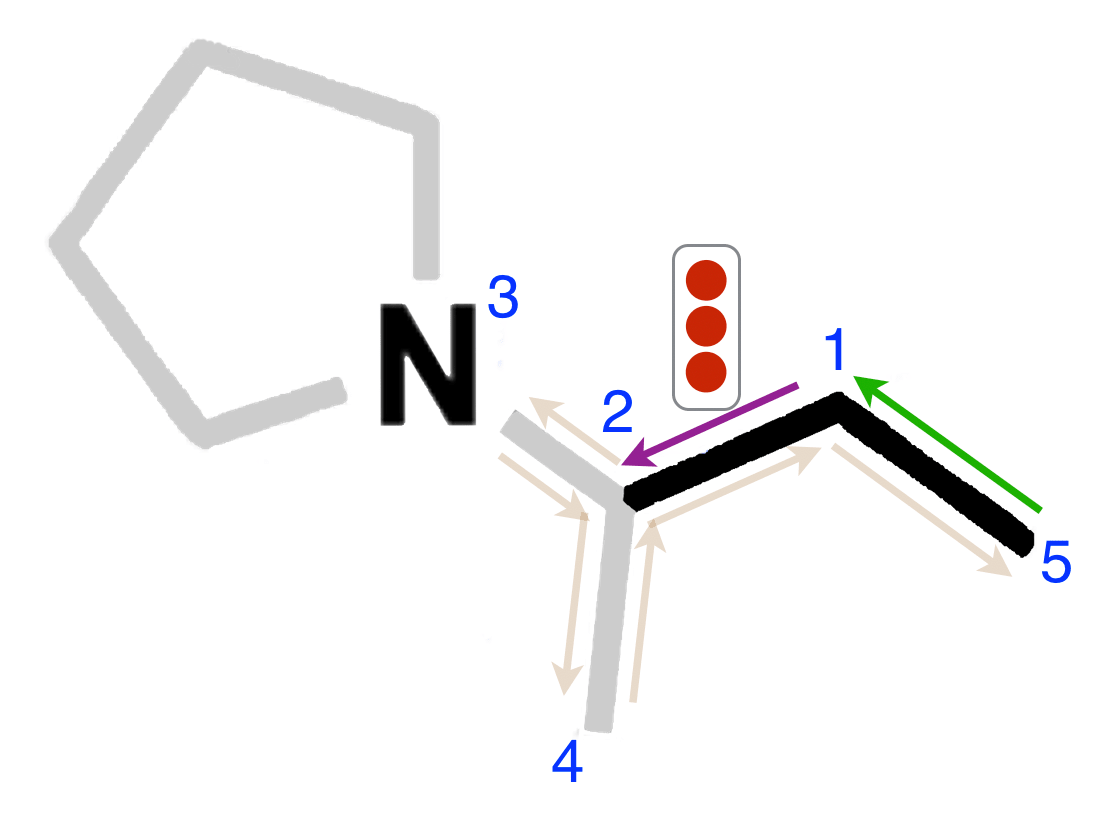}
            \caption{}
            \label{fig:bond_message_passingb}
        \end{subfigure}
    \end{minipage}
    \hfill
    \begin{subfigure}{.5\textwidth}
        \includegraphics[width=\linewidth]{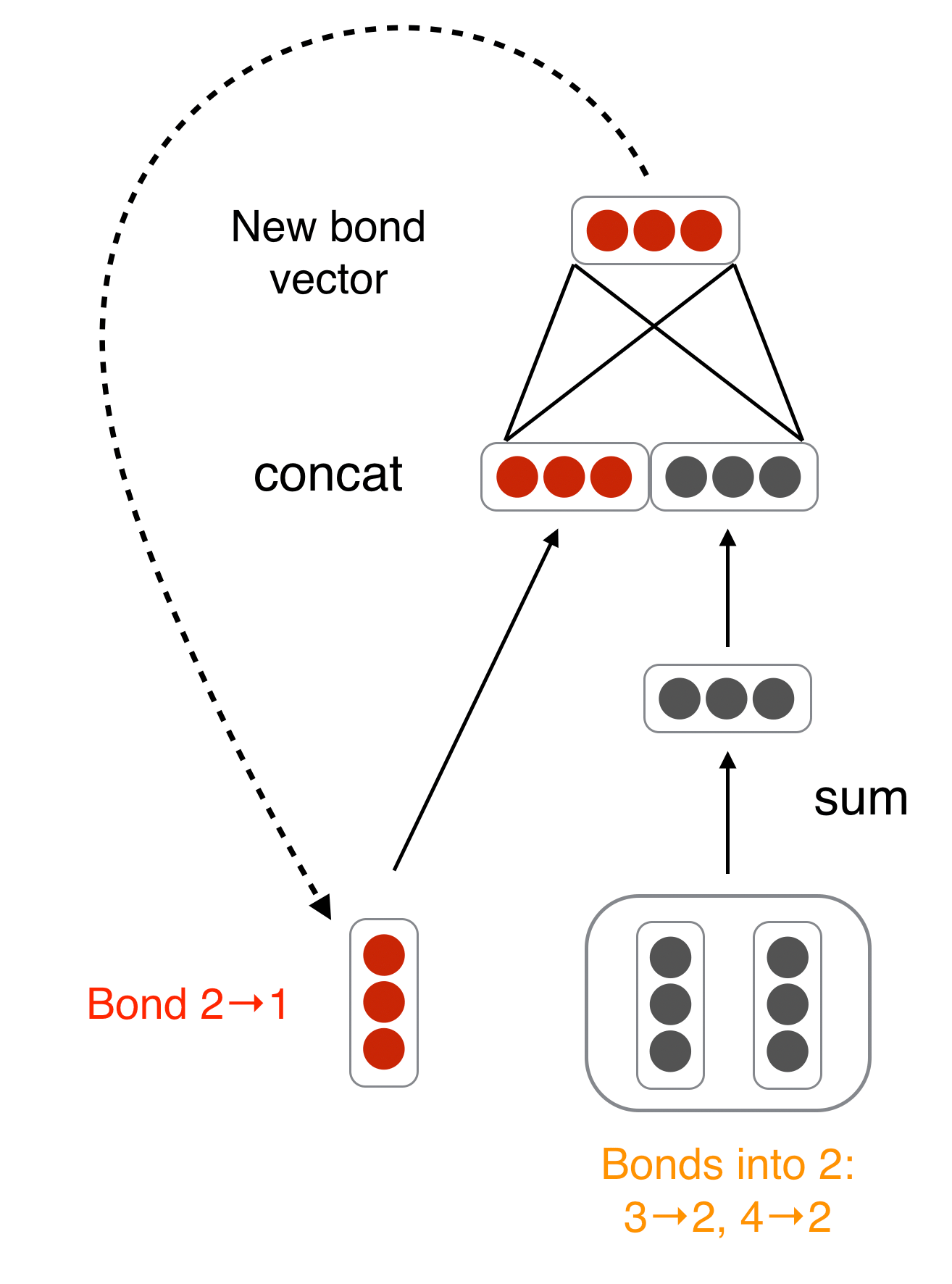}
        \caption{}
        \label{fig:bond_message_passingc}
    \end{subfigure}
    \caption{Illustration of bond-level message passing in our proposed D-MPNN. (a): Messages from the orange directed bonds are used to inform the update to the hidden state of the red directed bond. By contrast, in a traditional MPNN, messages are passed from atoms to atoms (for example atoms 1, 3, and 4 to atom 2) rather than from bonds to bonds. (b): Similarly, a message from the green bond informs the update to the hidden state of the purple directed bond. (c): Illustration of the update function to the hidden representation of the red directed bond from diagram (a).}
    \label{fig:bond_message_passing}
\end{figure}

The main difference between the Directed MPNN (D-MPNN)\cite{dai2016discriminative} and the generic MPNN described above is the nature of the messages sent during the message passing phase. Rather than using messages associated with vertices (atoms), D-MPNN uses messages associated with directed edges (bonds).
%wengong's modification
The motivation of this design is to prevent totters~\citep{mahe2004extensions}, that is, to avoid messages being passed along any path of the form $v_1v_2 \cdots v_n$ where $v_i=v_{i+2}$ for some $i$. Such excursions are likely to introduce noise into the graph representation. Using Figure~\ref{fig:bond_message_passing} as an illustration, in D-MPNN, the message $1\rightarrow2$ will only be propagated to nodes 3 and 4 in the next iteration, whereas in the original MPNN it will be sent to node 1 as well, creating an unnecessary loop in the message passing trajectory. Compared to the atom based message passing approach, this message passing procedure is more similar to belief propagation in probabilistic graphical models\cite{koller2009probabilistic}. We refer to \citeauthor{dai2016discriminative}\cite{dai2016discriminative} for futher discussion about the connection between D-MPNN and belief propagation.
%This aligns more closely with the belief propagation algorithms\cite{coughlan2009tutorial} that can be viewed as an inspiration to the message-passing paradigm for constructing fingerprints, and we hypothesize that the decreased symmetry from using directed bonds boosts performance as well. 

\FloatBarrier

The D-MPNN works as follows. The D-MPNN operates on hidden states $h_{vw}^t$ and messages $m_{vw}^t$ instead of on node based hidden states $h_v^t$ and messages $m_v^t$. Note that the direction of messages matters (i.e., $h_{vw}^t$ and $m_{vw}^t$ are distinct from $h_{wv}^t$ and $m_{wv}^t$). The corresponding message passing update equations are thus
\begin{align*}
    m_{vw}^{t+1} &= \sum_{k \in \{N(v) \setminus w\}} M_t(x_v, x_k, h_{kv}^t) \\
    h_{vw}^{t+1} &= U_t(h_{vw}^t, m_{vw}^{t+1}).
\end{align*}
Observe that message $m_{vw}^{t+1}$ does not depend on its reverse message $m_{wv}^t$ from the previous iteration.
Prior to the first step of message passing, we initialize edge hidden states with
\begin{equation*}
    h_{vw}^0 = \tau(W_i\ {\tt cat}(x_v, e_{vw}))
\end{equation*}
where $W_i \in \mathbb{R}^{h \times h_i}$ is a learned matrix, ${\tt cat}(x_v, e_{vw}) \in \mathbb{R}^{h_i}$ is the concatenation of the atom features $x_v$ for atom $v$ and the bond features $e_{vw}$ for bond $vw$, and $\tau$ is the ReLU activation function\cite{nair2010relu}.

We choose to use relatively simple message passing functions $M_t$ and edge update functions $U_t$. Specifically, we define $M_t(x_v, x_w, h_{vw}^t) = h_{vw}^t$ and we implement $U_t$ with the same neural network on every step,
\begin{equation*}
    U_t(h_{vw}^t, m_{vw}^{t+1}) = U(h_{vw}^t, m_{vw}^{t+1}) = \tau(h_{vw}^0 + W_m m_{vw}^{t+1})
\end{equation*}
where $W_m \in \mathbb{R}^{h \times h}$ is a learned matrix with hidden size $h$. Note that the addition of $h_{vw}^0$ on every step provides a skip connection to the original feature vector for that edge.

Finally, we return to an atom representation of the molecule by summing the incoming bond features according to
\begin{align*}
    m_v &= \sum_{k \in N(v)} h_{kv}^T \\
    h_v &= \tau(W_a {\tt cat}(x_v, m_v))
\end{align*}
where $W_a \in \mathbb{R}^{h \times h}$ is a learned matrix.

Altogether, the D-MPNN message passing phase operates according to
\begin{equation*}
    h_{vw}^0 = \tau(W_i\ {\tt cat}(x_v, e_{vw}))
\end{equation*}
followed by
\begin{align*}
    m_{vw}^{t+1} &= \sum_{k \in \{N(v) \setminus w\}} h_{kv}^t \\
    h_{vw}^{t+1} &= \tau(h_{vw}^0 + W_m m_{vw}^{t+1})
\end{align*}
for $t \in \{1, \dots, T\}$, followed by
\begin{align*}
    m_v &= \sum_{w \in N(v)} h_{vw}^T \\
    h_v &= \tau(W_a {\tt cat}(x_v, m_v)).
\end{align*}

The readout phase of the D-MPNN is the same as the readout phase of a generic MPNN. In our implementation of the readout function $R$, we first sum the atom hidden states to obtain a feature vector for the molecule
\begin{equation*}
    h = \sum_{v \in G} h_v.
\end{equation*}
Finally, we generate property predictions $\hat{y} = f(h)$ where $f(\cdot)$ is a feed-forward neural network.

\subsection{Initial Featurization}

Our model's initial atom and bond features are listed in Tables \ref{tab:atom_features} and \ref{tab:bond_features}, respectively. The D-MPNN's initial node features $x_v$ are simply the atom features for that node, while the D-MPNN's initial edge features $e_{vw}$ are the bond features for bond $vw$. All features are computed using the open-source package RDKit\cite{landrum2006rdkit}. 
% CWC: for stereochemistry of double bonds, e/z and cis/trans are presumably redundant
% CWC: also, to be pedantic, does it count as a one-hot encoding if it is a boolean value?

\FloatBarrier

% see https://arxiv.org/pdf/1603.00856.pdf page 7 for example table of atom features
\begin{table}[]
    \centering
    \begin{tabular}{|c|c|c|}
        \hline
        \textbf{Feature} & \textbf{Description} & \textbf{Size} \\
        \Xhline{2\arrayrulewidth}
        Atom type & Type of atom (ex. C, N, O), by atomic number. & 100 \\
        \# Bonds & Number of bonds the atom is involved in. & 6 \\
        Formal charge & Integer electronic charge assigned to atom. & 5 \\
        Chirality & Unspecified, tetrahedral CW/CCW, or other. & 4 \\
        \# Hs & Number of bonded Hydrogen atom. & 5 \\
        Hybridization & sp, sp\textsuperscript{2}, sp\textsuperscript{3}, sp\textsuperscript{3}d, or sp\textsuperscript{3}d\textsuperscript{2}. & 5 \\
        Aromaticity & Whether this atom is part of an aromatic system. & 1 \\
        Atomic mass & Mass of the atom, divided by 100. & 1 \\
        \hline
    \end{tabular}
    \caption{Atom Features. All features are one-hot encodings except for atomic mass, which is a real number scaled to be on the same order of magnitude.}
    \label{tab:atom_features}
\end{table}

\begin{table}[]
    \centering
    \begin{tabular}{|c|c|c|}
        \hline
        \textbf{Feature} & \textbf{Description} & \textbf{Size} \\
        \Xhline{2\arrayrulewidth}
        Bond type & Single, double, triple, or aromatic. & 4 \\
        Conjugated & Whether the bond is conjugated. & 1 \\
        In ring & Whether the bond is part of a ring. & 1 \\
        Stereo & None, any, E/Z or cis/trans. & 6 \\
        \hline
    \end{tabular}
    \caption{Bond Features. All features are one-hot encodings.}
    \label{tab:bond_features}
\end{table}

\subsection{D-MPNN with Features}

Next, we discuss further extensions and optimizations to improve performance. Although an MPNN should ideally be able to extract \emph{any} information about a molecule that might be relevant to predicting a given property, two limitations may prevent this in practice. First, many property prediction datasets are very small, i.e., on the order of only hundreds or thousands of molecules. With so little data, MPNNs are unable to learn to identify and extract all features of a molecule that might be relevant to property prediction, and they are susceptible to overfitting to artifacts in the data. Second, most MPNNs use fewer message passing steps than the diameter of the molecular graph, i.e. $T < {\tt diam}(G)$, meaning atoms that are a distance of greater than $T$ bonds apart will never receive messages about each other. This results in a molecular representation that is fundamentally local rather than global in nature, meaning the MPNN may struggle to predict properties that depend heavily on global features. 

In order to counter these limitations, we introduce a variant of the D-MPNN that incorporates 200 global molecular features that can be computed rapidly \textit{in silico} using RDKit. The neural network architecture requires that the features are appropriately scaled to prevent features with large ranges dominating smaller ranged features, as well as preventing issues where features in the training set are not drawn from the same sample distribution as features in the testing set. To prevent these issues, a large sample of molecules was used to fit cumulative density functions (CDFs) to all features. CDFs were used as opposed to simpler scaling algorithms mainly because CDFs have the useful property that each value has the same meaning: the percentage of the population observed below the raw feature value. Min-max scaling can be easily biased with outliers and Z-score scaling assumes a normal distribution which is most often not the case for chemical features, especially if they are based on counts.

\begin{figure}
    \centering
    \begin{subfigure}[b]{0.475\textwidth}
        \centering
        \includegraphics[width=\textwidth]{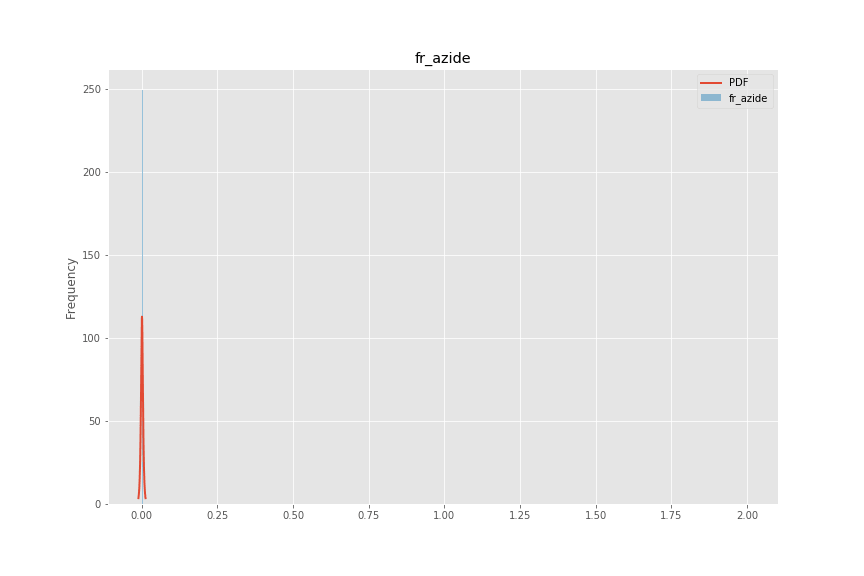}
        \caption{fr\_azide}
        \label{fig:cdfsa}
    \end{subfigure}
    \hfill
    \begin{subfigure}[b]{0.475\textwidth}  
        \centering 
        \includegraphics[width=\textwidth]{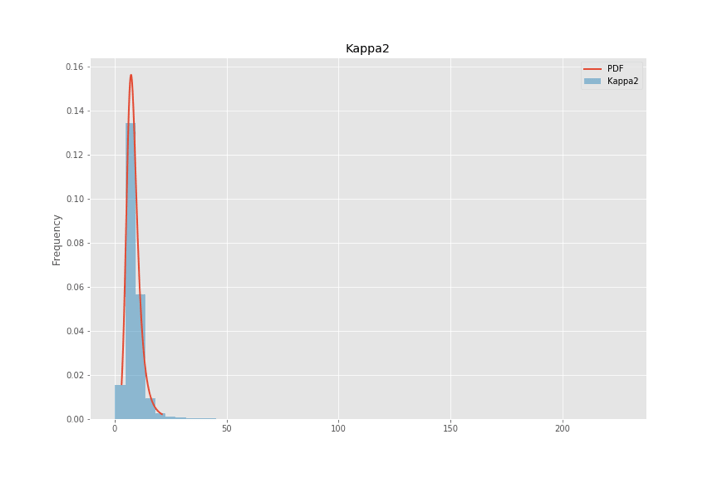}
        \caption{Kappa2}
        \label{fig:cdfsb}
    \end{subfigure}
    \vskip\baselineskip
    \begin{subfigure}[b]{0.475\textwidth}   
        \centering 
        \includegraphics[width=\textwidth]{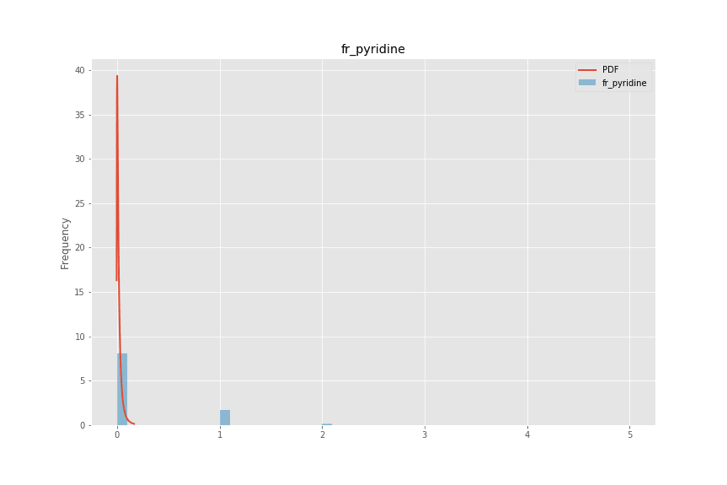}
        \caption{fr\_pyridine}
        \label{fig:cdfsc}
    \end{subfigure}
    \hfill
    \begin{subfigure}[b]{0.475\textwidth}   
        \centering 
        \includegraphics[width=\textwidth]{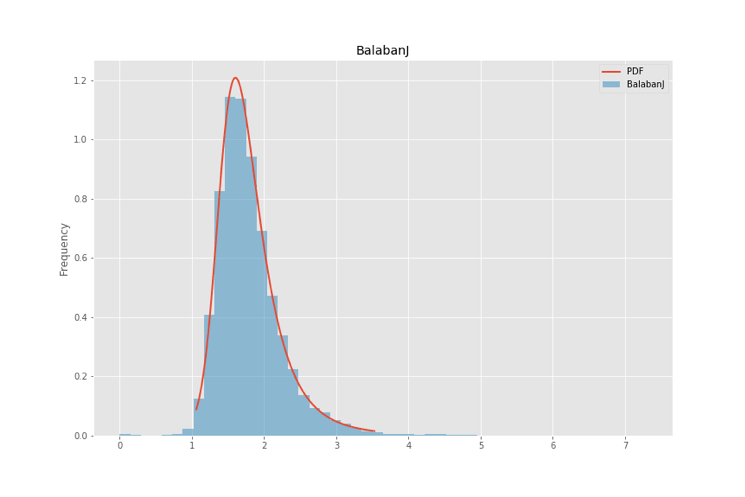}
        \caption{BalabanJ}
        \label{fig:cdfsd}
    \end{subfigure}
    \caption{Four example distributions fit to a random sample of 100,000 compounds used for biological screening in Novartis. Note that some distributions for discrete calculations, such as fr\_pyridine, are not fit especially well. This is an active area for improvement.}
    \label{fig:cdfs}
\end{figure}

The CDFs were fit to a sample of 100k compounds from the Novartis internal catalog using the distributions available in the scikit-learn package\cite{pedregosa2011scikit}, a sample of which can be seen in Figure \ref{fig:cdfs}. One could do a similar normalization using publicly available databases such as ZINC\cite{irwin2005zinc} and PubChem\cite{kim2015pubchem}. scikit-learn was used primarily due to the simplicity of fitting and the final application. However, more complicated techniques could be used in the future to fit to empirical CDFs, such as finding the best fit general logistic function, which has been shown to be successful for other biological datasets\cite{sartor2008lrpath}. No review was taken to remove odd distributions. For example, azides are hazardous and rarely used outside of a few specific reactions, as reflected in the fr\_azide distribution in Figure \ref{fig:cdfs}. As such, since the sample data was primarily used for chemical screening against biological targets, the distribution used here may not accurately reflect the distribution of reagents used for chemical synthesis. For the full list of calculated features, please refer to the Supporting Information. 

\FloatBarrier

% CWC: can you increase font sizes + label each plot with a, b, c, and d? 

% CWC: I don't think this next sentence is necessary
% Future work should include detecting ill-fitting distributions, such as for fr\_pyridine as seen in Figure \ref{fig:cdfs}, and using alternative methods, such as empirical distributions.

To incorporate these features, we modify the readout phase of the D-MPNN to apply the feed-forward neural network $f$ to the concatenation of the learned molecule feature vector $h$ and the computed global features $h_f$,
\begin{equation*}
    \hat{y} = f({\tt cat}(h, h_f)).
\end{equation*}
This is a very general method of incorporating external information and can be used with any MPNN and any computed features or descriptors.

\subsection{Hyperparameter Optimization}

The performance of MPNNs, like most neural networks, can depend greatly on the settings of the various model hyperparameters, such as the hidden size of the neural network layers. Thus to maximize performance, we perform hyperparameter optimization via Bayesian Optimization\cite{shahriari2016bayes} using the {\tt Hyperopt}\cite{hyperopt} Python package. We specifically optimize our model's depth (number of message-passing steps), hidden size (size of bond message vectors), number of feed-forward network layers, and dropout probability.

\subsection{Ensembling}

A common technique in machine learning for improving model performance is ensembling, where the predictions of multiple independently trained models are combined to produce a more accurate prediction\cite{dietterich2000ensemble}. We apply this technique by training several copies of our model, each initialized with different random weights, and then averaging the predictions of these models (each with equal weight) to generate an ensemble prediction.

Since prior work did not report performance using ensembling, all direct comparisons we make to prior work use a single D-MPNN model for a fair comparison. However, we also report results using an ensemble to illustrate the maximum possible performance of our model architecture.

\subsection{Implementation}

We implement our model using the PyTorch\cite{paszke2017pytorch} deep learning framework. All code for the D-MPNN and its variants is available in our GitHub repository \cite{chemprop_code}. Code for computing and using the RDKit feature CDFs is available in the \texttt{Descriptastorus} package\cite{descriptastorus}. Additionally, a web demonstration of our model's predictive capability on public datasets is available online\cite{chemprop_website}. 

\section{Experiments}

\subsection{Data}

We test our model on 19 publicly available datasets from \citeauthor{Wu_2018}\cite{Wu_2018} and \citeauthor{mayr2018chembl}\cite{mayr2018chembl}. These datasets range in size from less than 200 molecules to over 450,000 molecules. They include a wide range of regression and classification targets spanning quantum mechanics, physical chemistry, biophysics, and physiology. Detailed descriptions are provided in Table \ref{tab:dataset_descriptions}.

\FloatBarrier

\begin{table}[]
\resizebox{\textwidth}{!}{
\centering
\begin{tabular}{|c|c|l|}
\hline
\textbf{Dataset} & \textbf{Category} & \textbf{Description} \\
\Xhline{2\arrayrulewidth}
QM7, QM8, QM9 & Quantum Mechanics & Computer-generated quantum mechanical properties \\
ESOL & Physical Chemistry & Water solubility \\
FreeSolv & Physical Chemistry & Hydration free energy in water \\
Lipophilicity & Physical Chemistry & Octanol/water distribution coefficients \\
PDBbind & Biophysics & Protein binding affinity \\
\Xhline{2\arrayrulewidth}
PCBA & Biophysics & Assorted biological assays \\
MUV & Biophysics & Assorted biological assays \\
HIV & Biophysics & Inhibition of HIV replication \\
BACE & Biophysics & Inhibition of human $\beta$-secretase $1$ \\
BBBP & Physiology & Ability to penetrate the blood-brain-barrier \\
Tox21 & Physiology & Toxicity \\
ToxCast & Physiology & Toxicity \\
SIDER & Physiology & Side effects of drugs \\
ClinTox & Physiology & Toxicity \\
ChEMBL & Physiology & Biological assays \\
\hline
\end{tabular}}
\caption{Descriptions of the public datasets used in this paper.}
\label{tab:dataset_descriptions}
\end{table}

Summary statistics for all the datasets are provided in Table \ref{tab:dataset_stats}, and further details on the datasets are available in \citeauthor{Wu_2018}\cite{Wu_2018}, with the exception of the ChEMBL dataset which is described in \citeauthor{mayr2018chembl}\cite{mayr2018chembl}. Additional information on the class balance of the classification datasets is provided in the Supporting Information. Although most classification datasets are reasonably balanced, the MUV dataset is particularly unbalanced, with only 0.2\% of molecules classified as positive. This makes our model unstable, leading to the wide variation in performance on this dataset in the subsequent sections.

It is worth noting that for some datasets, the number of compounds in Table \ref{tab:dataset_stats} does not precisely match the numbers from \citeauthor{Wu_2018}\cite{Wu_2018}. This is because \citeauthor{Wu_2018}\cite{Wu_2018} included duplicate molecules in that count while we count the unique number of molecules. Additionally, we left out one or two molecules which could not be processed by RDKit\cite{landrum2006rdkit}. However, the impact of removing these molecules is negligible on overall model performance. Furthermore, we have fewer molecules in QM7 because we used SMILES strings generated by \citeauthor{Wu_2018}\cite{Wu_2018} from the original 3D coordinates in the dataset, but the SMILES conversion process failed for $\sim 300$ molecules. For this reason, we do not directly compare our model's performance on QM7 to the QM7 performance numbers reported by \citeauthor{Wu_2018}\cite{Wu_2018}.

\FloatBarrier

\begin{table}[]
\resizebox{\textwidth}{!}{
\centering
\begin{tabular}{|c|c|c|c|c|}
\hline
\textbf{Dataset}           & \textbf{\# Tasks} & \textbf{Task Type}      & \textbf{\# Compounds} & \textbf{Metric}  \\
\Xhline{2\arrayrulewidth}
QM7               & 1        & Regression     & 6,830         & MAE     \\
QM8               & 12       & Regression     & 21,786        & MAE     \\
QM9               & 12       & Regression     & 133,885       & MAE     \\
ESOL              & 1        & Regression     & 1,128         & RMSE    \\
FreeSolv          & 1        & Regression     & 642           & RMSE    \\
Lipophilicity     & 1        & Regression     & 4,200         & RMSE    \\
PDBbind-F         & 1        & Regression     & 9,880         & RMSE    \\
PDBbind-C         & 1        & Regression     & 168           & RMSE    \\
PDBbind-R         & 1        & Regression     & 3,040         & RMSE    \\
\Xhline{2\arrayrulewidth}
PCBA              & 128      & Classification & 437,929       & PRC-AUC \\
MUV               & 17       & Classification & 93,087        & PRC-AUC \\
HIV               & 1        & Classification & 41,127        & ROC-AUC \\
BACE              & 1        & Classification & 1,513         & ROC-AUC \\
BBBP              & 1        & Classification & 2,039         & ROC-AUC \\
Tox21             & 12       & Classification & 7,831         & ROC-AUC \\
ToxCast           & 617      & Classification & 8,576         & ROC-AUC \\
SIDER             & 27       & Classification & 1,427         & ROC-AUC \\
ClinTox           & 2        & Classification & 1,478         & ROC-AUC \\
ChEMBL            & 1,310     & Classification & 456,331      & ROC-AUC \\
\hline
\end{tabular}}
\caption{Summary statistics of the public datasets used in this paper. Note: PDBbind-F, PDBbind-C, and PDBbind-R refer to the full, core, and refined PDBbind datasets from \citeauthor{Wu_2018}\cite{Wu_2018}.}
\label{tab:dataset_stats}
\end{table}

\subsection{Experimental Procedure}

\paragraph{Cross-Validation and Hyperparameter Optimization.}

Since many of the datasets are very small (two thousand molecules or fewer), we use a cross-validation approach to decrease noise in the results both while optimizing the hyperparameters and while determining final performance numbers. For consistency, we maintain the same approach for all of our datasets. Specifically, for each dataset, we use 20 iterations of Bayesian optimization on 10 randomly-seeded 80:10:10 data splits to determine the best hyperparameters, selecting hyperparameters based on validation set performance. We then evaluate the model by retraining using the optimal hyperparameters and checking performance on the test set. Due to computational cost, we only use 3 splits for HIV, QM9, MUV, PCBA, and ChEMBL. When we run the best model from \citeauthor{mayr2018chembl}\cite{mayr2018chembl} for comparative purposes, we optimize their model's hyperparameters with the same splits, using their original hyperparameter optimization script.

\paragraph{Split Type.} We evaluate all models on random and scaffold-based splits as well as on the original splits from \citeauthor{Wu_2018}\cite{Wu_2018} and \citeauthor{mayr2018chembl}\cite{mayr2018chembl}. The one exception is the model of \citeauthor{mayr2018chembl}\cite{mayr2018chembl}, which we only ran on scaffold-based splits, due to the large computational cost of optimizing their model. Results on scaffold-based splits are reported below while results on random splits are presented in the Supporting Information.

Our scaffold split is similar to that of \citeauthor{Wu_2018}\cite{Wu_2018}. Molecules are partitioned into bins based on their Murcko scaffold calculated by RDKit\cite{landrum2006rdkit}. Any bins larger than half of the desired test set size are placed into the training set, in order to guarantee the scaffold diversity of the validation and test sets. All remaining bins are placed randomly into the training, validation, and test sets until each set has reached its desired size. As this latter process involves randomly placing scaffolds into bins, we are able to generate several different scaffold splits for evaluation. 

None of our splits on classification datasets are stratified; we do not enforce class balance. Compared to random splits, the scaffold splits have more class imbalance on average, but are not excessively imbalanced; we analyze this class balance quantitatively in the Additional Dataset Statistics section of the Supporting Information.

Compared to a random split, a scaffold split is a more challenging and realistic evaluation setting as shown in Figures \ref{fig:amgen_split_type} and \ref{fig:pdbbind_split_type}. This allows us to use a scaffold split as a proxy for the chronological split present in real-world property prediction data, where one trains a model on past data to make predictions on future data, although chronological splits are still preferred when available. However, as chronological information is not available for most public datasets, we use a scaffold-based split for all evaluations except for our direct comparison with the MoleculeNet models from \citeauthor{Wu_2018}\cite{Wu_2018}, for which we use their original data splits.

\paragraph{Baselines.} We compare our model to the following baselines:

\begin{itemize}
    \item The best model for each dataset from MoleculeNet by \citeauthor{Wu_2018}\cite{Wu_2018}
    \item The best model from \citeauthor{mayr2018chembl}\cite{mayr2018chembl}, a feed-forward neural network on a concatenation of assorted expert-designed molecular fingerprints.
    \item Random forest on binary Morgan fingerprints.
    \item Feed-forward network (FFN) on binary Morgan fingerprints using the same FFN architecture that our D-MPNN uses during its readout phase.
    \item FFN on count-based Morgan fingerprints.
    \item FFN on RDKit-calculated descriptors.
\end{itemize}

The models in MoleculeNet by \citeauthor{Wu_2018}\cite{Wu_2018} include MPNN\cite{gilmer2017neural}, Weave\cite{kearnes2016molecular}, GraphConv, kernel ridge regression, gradient boosting\cite{friedman2001greedy}, random forest\cite{breiman2001random}, logistic regression\cite{cramer2010logreg}, directed acyclic graph models\cite{lusci2013deep}, support vector machines\cite{vapnik1995svm}, Deep Tensor Neural Networks\cite{schutt2017quantum}, multitask networks\cite{ma2015deep}, bypass networks\cite{ramsundar2017multitask}, influence relevance voting\cite{swamidass2009influence}, and/or ANI-1\cite{smith2017ani}, depending on the dataset. Full details can be found in \citeauthor{Wu_2018}\cite{Wu_2018}. For the feed-forward network model from \citeauthor{mayr2018chembl}\cite{mayr2018chembl}, we modified the authors' original code with their guidance in order to run their code on all of the datasets, not just on the ChEMBL dataset they experimented with. We tuned learning rates and hidden dimensions in addition to the extensive hyperparameter search already present in their code.

\section{Results and Discussion}

In the following sections, we analyze the performance of our model on both public and proprietary datasets. Specifically, we aim to answer the following questions:
\begin{enumerate}
    \item How does our model perform on both public and proprietary datasets compared to public benchmarks, and how close are we to the upper bound on performance represented by experimental reproducibility?
    \item How should we be splitting our data, and how does the method of splitting affect our evaluation of the model's generalization performance?
    \item What are the key elements of our model, and how can we maximize its performance? % CWC: key elements of our model that account for its performance?
\end{enumerate}

In the following sections, all results using root-mean-square error (RMSE) or mean absolute error (MAE) are displayed as plots showing change relative to a baseline model rather than showing absolute performance numbers. This is because the scale of the errors can differ drastically between datasets. All results using $R^2$, area under the receiver operating characteristic curve (ROC-AUC), or area under the precision recall curve (PRC-AUC) are displayed as plots showing the actual values. For RMSE and MAE, lower is better, while for $R^2$, ROC-AUC, and PRC-AUC, higher is better. Table \ref{tab:dataset_stats} indicates the metric used for each dataset. Tables showing the exact performance numbers for all experiments can be found in the Supporting Information. Note that the error bars on all plots show the standard error of the mean across multiple runs, where standard error is defined as the standard deviation divided by the square root of the number of runs.

We evaluate statistical significance using two statistical tests: a one-sided Wilcoxon signed-rank test and a one-sided Welch's t-test. While the Wilcoxon test is stronger, as it is a paired test comparing performance molecule-by-molecule, it requires knowing per-molecule predictions, which we do not have easy access to for the models from MoleculeNet\cite{Wu_2018} and \citeauthor{mayr2018chembl}\cite{mayr2018chembl}. Furthermore, comparisons between data split types inherently involves comparing performance on different test molecules, meaning a per-molecule test is not possible. Therefore, for these comparison we use the weaker Welch's t-test and for all other comparisons we use the Wilcoxon test. When using the Wilcoxon test for regression datasets, we directly compare test errors molecule-by-molecule. For the classification datasets, we divide all the test molecules into 30 equal parts, compute AUC on each part, and then use the Wilcoxon test on these AUC values. This subdivision of the test molecules into 30 parts gives the Wilcoxon test more strength than evaluating directly on the original 3 or 10 test cross-validation folds while still keeping each part large enough to result in a meaningful AUC computation. We define statistical significance as p-value less than 0.05.

Additionally, note that in all figures and tables, ``D-MPNN'' refers to the base D-MPNN model, ``D-MPNN Features'' refers to the D-MPNN with RDKit features, ``D-MPNN Optimized'' refers to the D-MPNN with RDKit features and optimized hyperparameters, and ``D-MPNN Ensemble'' refers to an ensemble of five D-MPNNs with RDKit features and optimized hyperparameters.

\subsection{Comparison to Baselines}

After optimizing our model, we compare our best single (non-ensembled) model on each dataset against models from prior work.

\subsubsection{Comparison to MoleculeNet}

\begin{figure}
\centering
\begin{subfigure}{\textwidth}
  \centering
  \includegraphics[width=\linewidth]{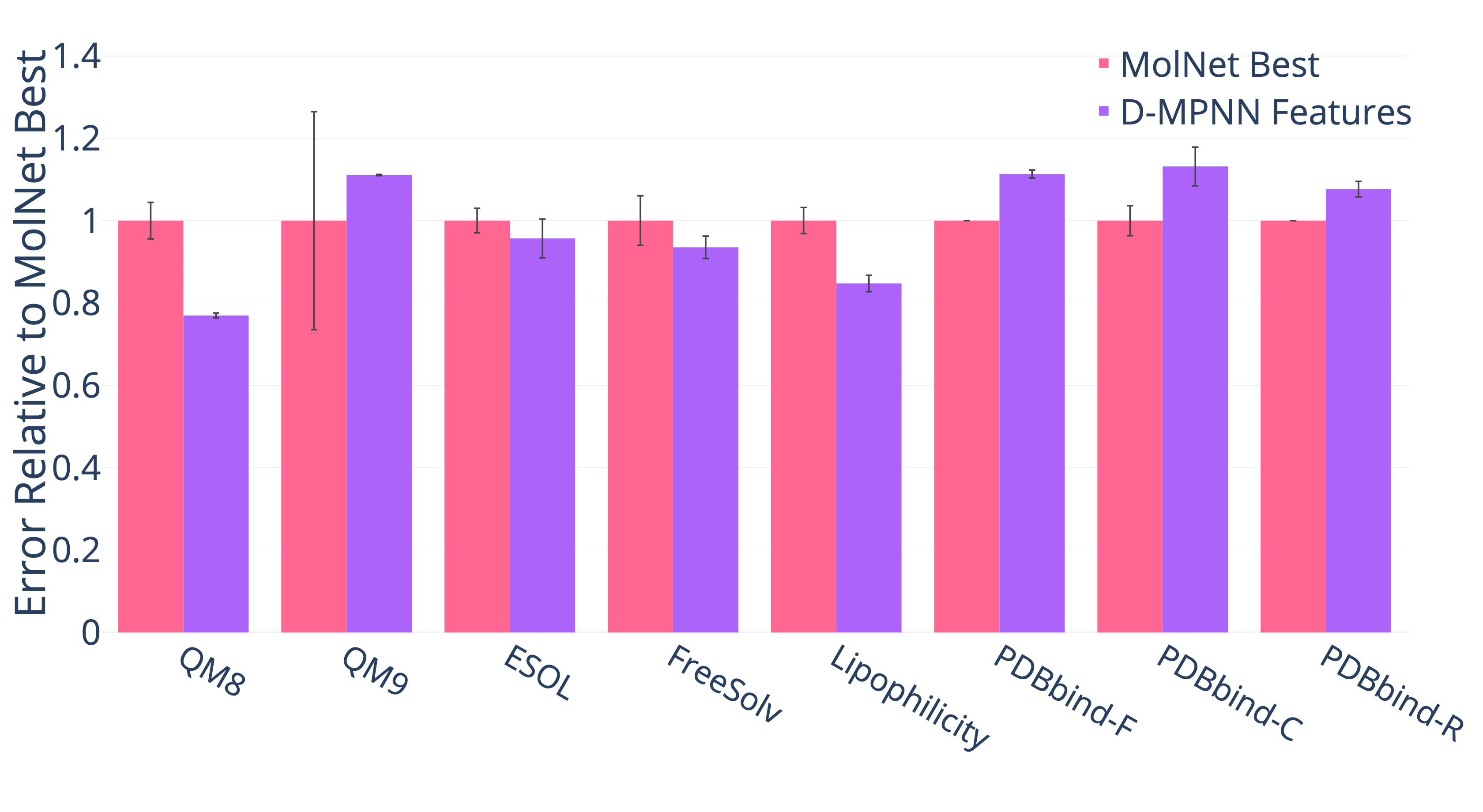}
  \caption{Regression Datasets (lower = better).}
  \label{fig:comparison_to_molneta}
\end{subfigure}%
\vskip\baselineskip
\begin{subfigure}{\textwidth}
  \centering
  \includegraphics[width=\linewidth]{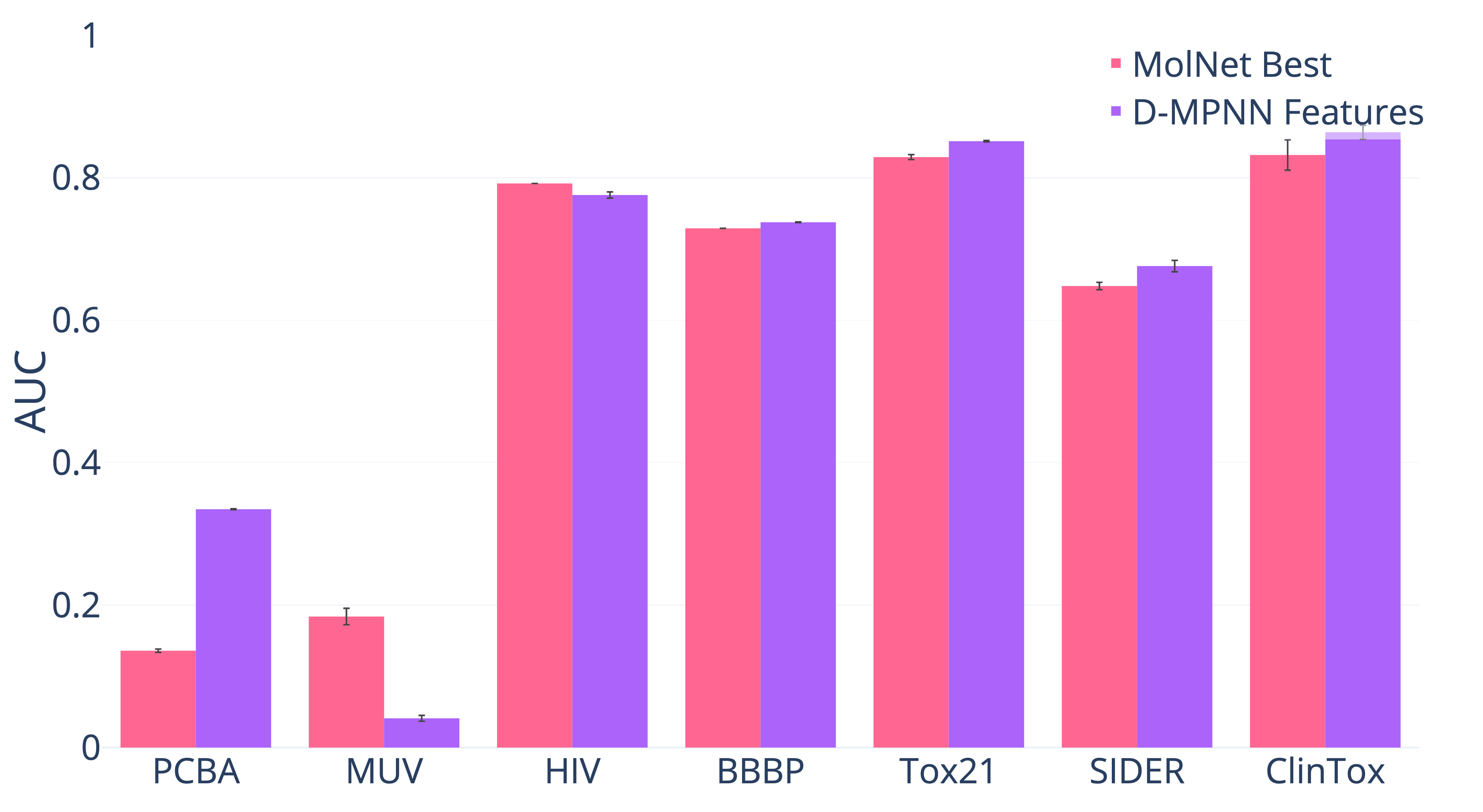}
  \caption{Classification Datasets (higher = better).}
  \label{fig:comparison_to_molnetb}
\end{subfigure}
\caption{Comparison of our D-MPNN with features to the best models from \citeauthor{Wu_2018}\cite{Wu_2018}.}
\label{fig:comparison_to_molnet}
\end{figure}

We first compare our D-MPNN to the best model from MoleculeNet\cite{Wu_2018,Ramsundar-et-al-2019} on the same datasets and splits on which \citeauthor{Wu_2018}\cite{Wu_2018} evaluate their models. We were unable to reproduce their original data splits on BACE, Toxcast, and QM7, but we have evaluated our model against their original splits on all of the other datasets. The splits are a mix of random, scaffold, and time splits, as indicated in Figure \ref{fig:comparison_to_molnet}.

Overall on the 10 datasets where the MoleculeNet models only use 2D information, i.e. all datasets except the QM and PDBbind datasets, our D-MPNN is significantly better than the best MoleculeNet models on 5 datasets, is not significantly different on 3 datasets, and is significantly worse on 2 datasets. This indicates that D-MPNN tends to outperform even the best MoleculeNet models, with the added benefit that the D-MPNN model architecture is the same for every dataset while the best MoleculeNet model architecture differs between datasets.

\FloatBarrier

Furthermore, we note that there are two cases in which our D-MPNN may underperform. The first is the MUV dataset, which is large but extremely imbalanced; only $0.2\%$ of samples are labeled as positives. \citeauthor{Wu_2018}\cite{Wu_2018} also encountered great difficulty with this extreme class imbalance when experimenting with the MUV dataset; all other datasets we experiment on contain at least $1$\% positives (see the Supporting Information for full class balance information). The second exception is when there is auxiliary 3D information available, as in the three variants of the PDBbind dataset and in QM9. The current iteration of our D-MPNN does not use 3D coordinate information, and we leave this extension to future work. Thus it is unsurprising that our D-MPNN model underperforms models using 3D information on a protein binding affinity prediction task such as PDBbind, where 3D structure is key. Nevertheless, our D-MPNN model outperforms the best graph-based method in MoleculeNet on PDBbind and QM9. Moreover, we note that on another dataset that provides 3D coordinate information, QM8, our model outperforms the best model in MoleculeNet with or without 3D coordinates. 

\subsubsection{Comparison to \citeauthor{mayr2018chembl}\cite{mayr2018chembl}}

\begin{figure}
\centering
\begin{subfigure}{\textwidth}
  \centering
  \includegraphics[width=\linewidth]{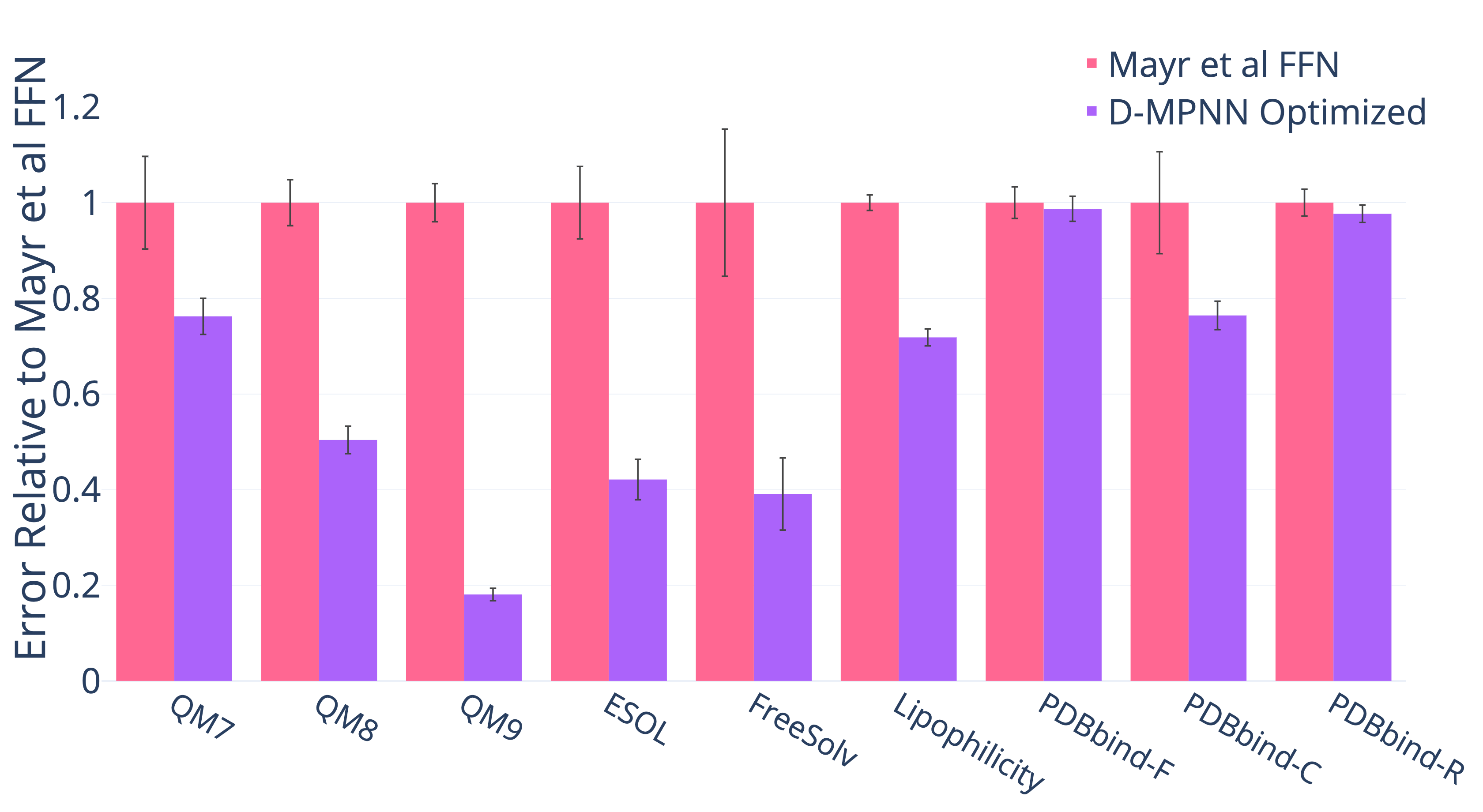}
  \caption{Regression Datasets (lower = better).}
  \label{fig:comparison_to_mayra}
\end{subfigure}%
\vskip\baselineskip
\begin{subfigure}{\textwidth}
  \centering
  \includegraphics[width=\linewidth]{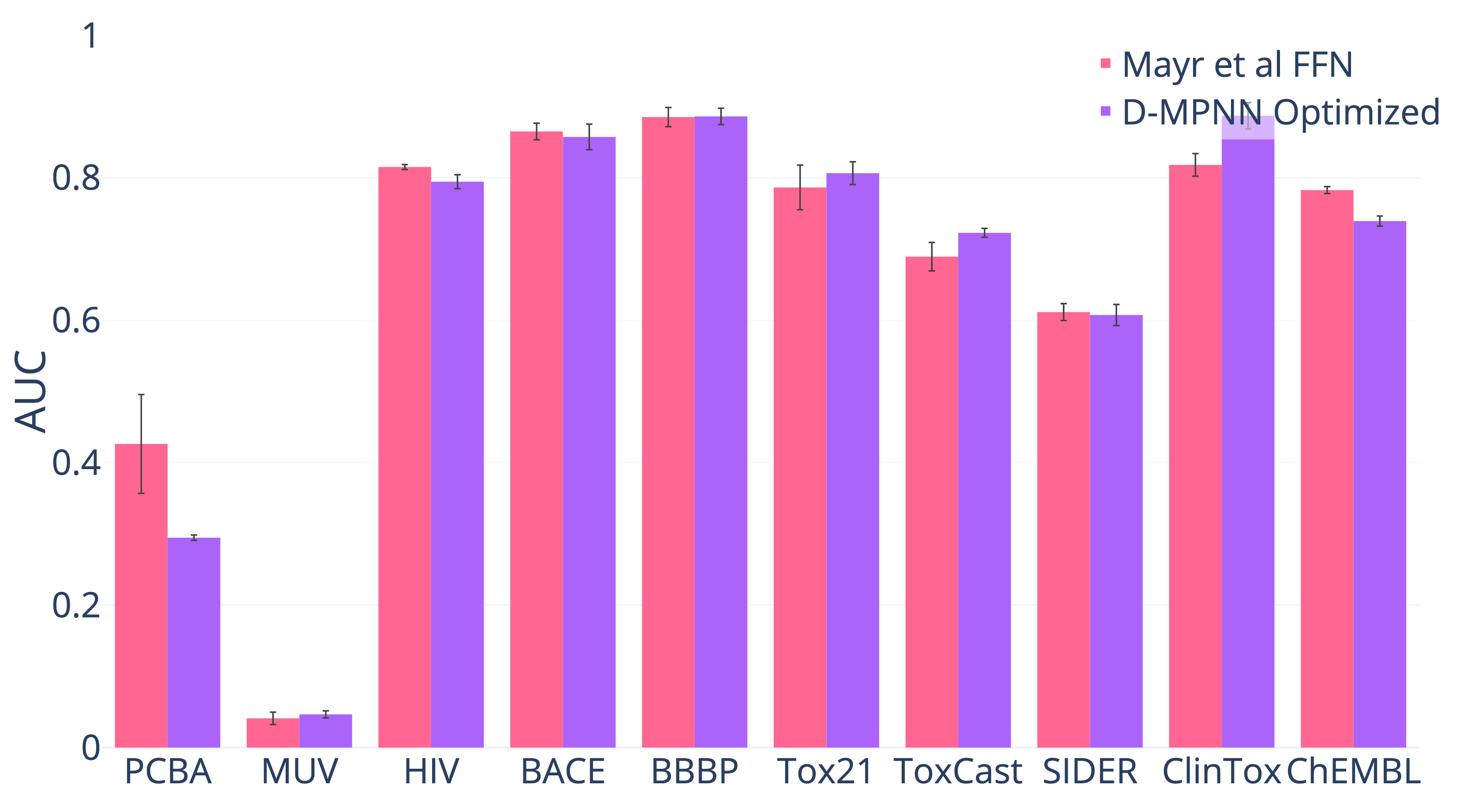}
  \caption{Classification Datasets (higher = better).}
  \label{fig:comparison_to_mayrb}
\end{subfigure}
\caption{Comparison of our best single model (i.e. optimized hyperparameters and RDKit features) to the model from \citeauthor{mayr2018chembl}.}
\label{fig:comparison_to_mayr}
\end{figure}

In addition, we compare D-MPNN to the baseline from \citeauthor{mayr2018chembl}\cite{mayr2018chembl} in Figure \ref{fig:comparison_to_mayr}. We reproduced the features from their best model on each dataset using their scripts or equivalent packages\cite{lsc_experiments}. We then ran their code and hyperparameter optimization directly on the classification datasets, and we modified their code to run on regression datasets with the authors' guidance\cite{lsc_experiments}. On most classification datasets, we obtain similar performance to \citeauthor{mayr2018chembl}\cite{mayr2018chembl}. On regression datasets, the baseline from \citeauthor{mayr2018chembl}\cite{mayr2018chembl} performs poorly in comparison, despite extensive tuning. We hypothesize that this poor performance on regression in comparison to classification is the result of a large number of binary input features to the output feed-forward network; this hypothesis is supported by the similarly poor performance of our Morgan fingerprint FFN baseline. In addition, their method does not employ early stopping based on validation set performance and therefore may overfit to the training data in some cases; this may be the source of some numerical instability. 

Overall, our D-MPNN is significantly better than the \citeauthor{mayr2018chembl}\cite{mayr2018chembl} model on 8 datasets, is not significantly different on 10 datasets, and is significantly worse on 1 dataset. This indicates that D-MPNN generally outperforms the \citeauthor{mayr2018chembl}\cite{mayr2018chembl} model, especially on regression datasets.

\FloatBarrier

%We note that while our D-MPNN outperforms \citeauthor{mayr2018chembl}'s\cite{mayr2018chembl} method on our scaffold split of the ChEMBL dataset, the reverse is true on \citeauthor{mayr2018chembl}'s\cite{mayr2018chembl} original splits, which are also scaffold-based but using a different methodology. %Interestingly, while our D-MPNN achieves a lower AUC on \citeauthor{mayr2018chembl}'s\cite{mayr2018chembl} original splits than on our own splits, \citeauthor{mayr2018chembl}'s\cite{mayr2018chembl} model achieves a lower AUC on our scaffold splits than on their original splits. %In this paper, we argue for the use of our scaffold splits, as they are empirically justified by a comparison to chronological splits on both public and industry datasets in Figures \ref{fig:amgen_split_type} and \ref{fig:pdbbind_split_type}.

% In addition, we observe that many of the targets in the ChEMBL dataset contain very few labels (often in the low hundreds), thus disadvantaging models that learn their own molecular representations from the data. We analyze this phenomenon more closely in Figure \ref{fig:dataset_size}. 

\subsubsection{Out-of-the-Box Comparison of D-MPNN to Other Baselines}

For our final baseline comparison, we evaluate our model's performance ``out-of-the-box,'' i.e. using all the default settings (hidden size = 300, depth = 3, number of feed-forward layers = 2, dropout = 0) without any hyperparameter optimization and without any additional features. For this comparison, we compare to a number of simple baseline models that use computed fingerprints or descriptors: 
\begin{enumerate}
    \item Random forest (RF) with 500 trees run on Morgan (ECFP) fingerprints using radius 2 and hashing to a bit vector of size 2048.
    \item Feed-forward network (FFN) on Morgan fingerprints.
    \item FFN on Morgan fingerprints which use substructure counts instead of bits.
    \item FFN on RDKit descriptors.
\end{enumerate}
The parameters of the simple baseline models are also out-of-the-box defaults. We make this comparison in order to demonstrate the strong out-of-the-box performance of our model across a wide variety of datasets. Finally, we include the performance of the automatically optimized version of our model as a reference. 

\begin{figure}
\centering
\begin{subfigure}{\textwidth}
  \centering
  \includegraphics[width=\linewidth]{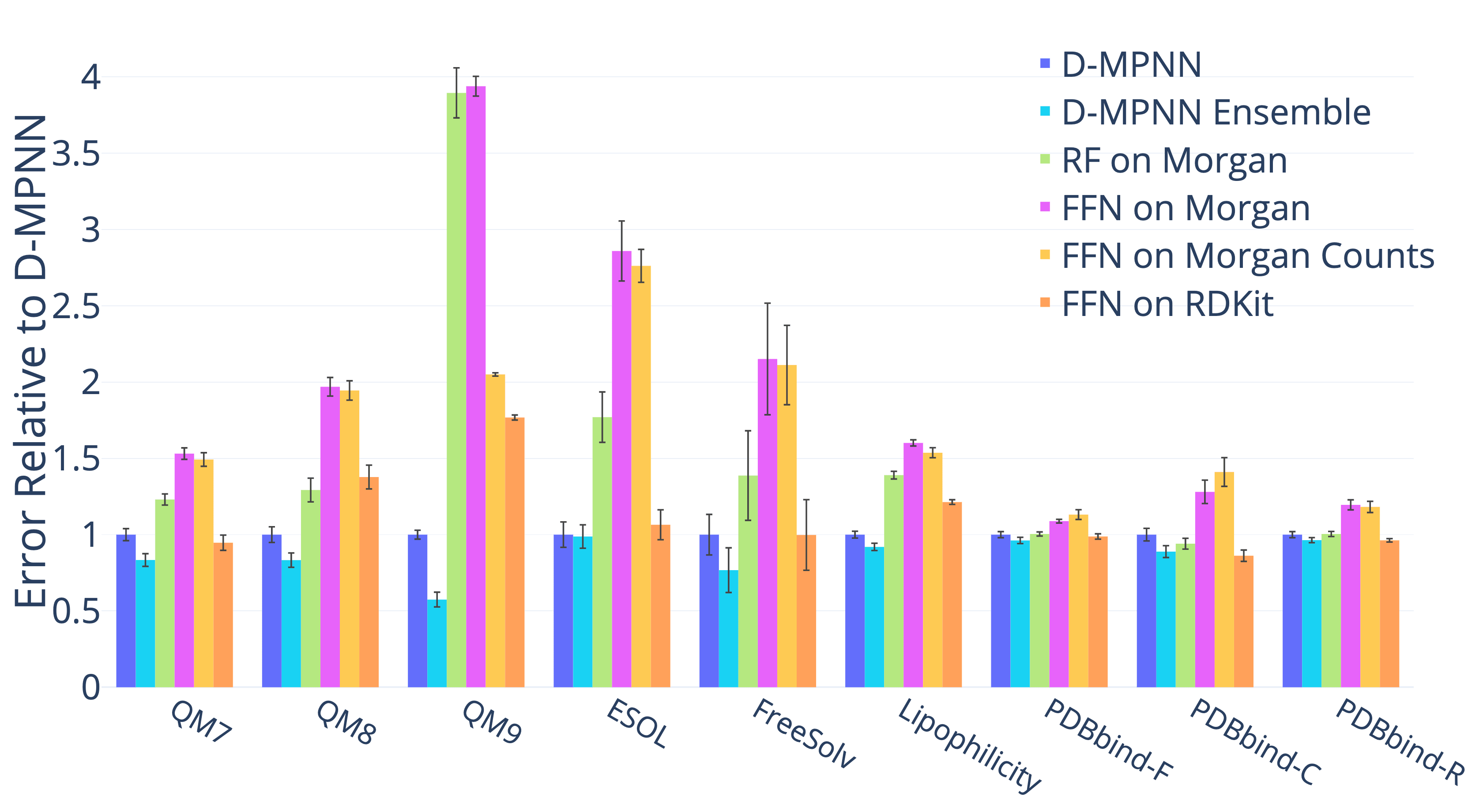}
  \caption{Regression Datasets (lower = better).}
  \label{fig:baselinesa}
\end{subfigure}%
\vskip\baselineskip
\begin{subfigure}{\textwidth}
  \centering
  \includegraphics[width=\linewidth]{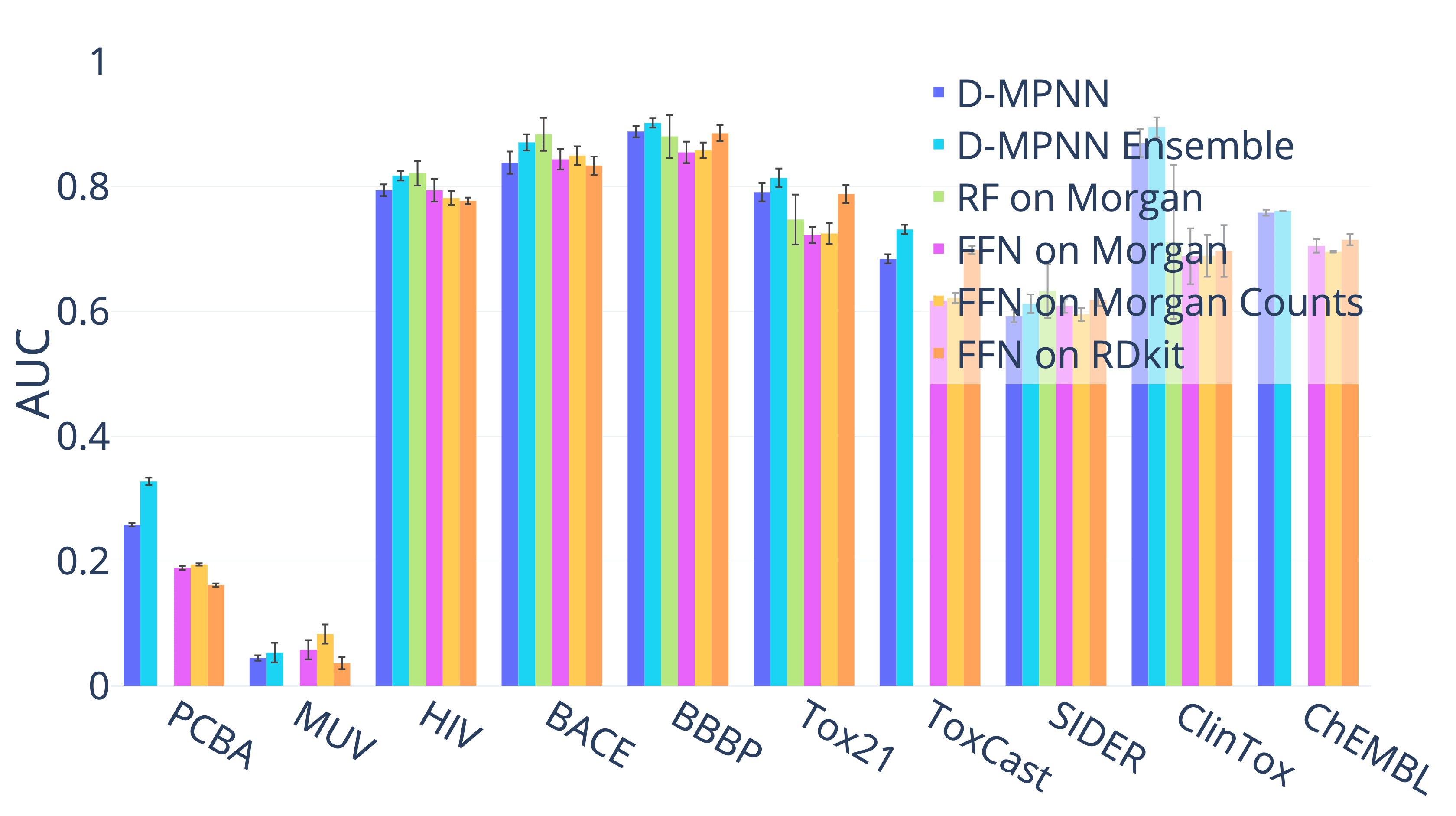}
  \caption{Classification Datasets (higher = better).}
  \label{fig:baselinesb}
\end{subfigure}
\caption{Comparison of our unoptimized D-MPNN against several baseline models. We omitted the random forest baseline on PCBA, MUV, Toxcast, and ChEMBL due to large computational cost. The D-MPNN matches or outperforms all baselines on 11 of the 19 datasets.}
\label{fig:baselines}
\end{figure}

Figure \ref{fig:baselines} shows that even without optimization, our D-MPNN provides an excellent starting point on a wide variety of datasets and targets, though it can be improved further with proper optimization.

\FloatBarrier

% CWC: same figure comments

\subsection{Proprietary Datasets}

We also ran our model on several private industry datasets, verifying that our model's strong performance on public datasets translates to real-world industrial datasets. 

\subsubsection{Amgen}

We ran our model along with \citeauthor{mayr2018chembl}'s\cite{mayr2018chembl} model and our simple baselines on four internal Amgen regression datasets. The datasets are as follows. 

\begin{enumerate}
    \item Rat plasma protein binding free fraction (rPPB).
    \item Solubility in 0.01 M hydrochloric acid solution, pH 7.4 phosphate buffer solution, and simulated intestinal fluid (Sol HCL, Sol PBS, and Sol SIF respectively).
    \item Rat liver microsomes intrinsic clearance (RLM).
    \item Human pregnane X receptor \% activation at 2uM and 10uM (hPXR).
\end{enumerate}

In addition, we binarized the hPXR dataset according to Amgen's recommendations in order to evaluate on a classification dataset. Details of the datasets are shown in Table \ref{tab:amgen_datasets}. Throughout the following, note that rPPB is in logit while Sol and RLM are in $\log_{10}$.

\FloatBarrier

\begin{table}[]
\resizebox{\textwidth}{!}{
\centering
\begin{tabular}{|c|c|c|c|c|c|}
\hline
\textbf{Category}           & \textbf{Dataset}     & \textbf{\# Tasks} & \textbf{Task Type}      & \textbf{\# Compounds} & \textbf{Metric}  \\
\Xhline{2\arrayrulewidth}
ADME               & rPPB        & 1        & Regression     & 1,441         & RMSE    \\
Physical Chemistry & Solubility  & 3        & Regression     & 18,007        & RMSE    \\
ADME               & RLM         & 1        & Regression     & 64,862        & RMSE    \\
ADME               & hPXR        & 2        & Regression     & 22,188        & RMSE    \\
\Xhline{2\arrayrulewidth}
ADME               & hPXR (class) & 2        & Classification & 22,188        & ROC-AUC \\
\hline
\end{tabular}}
\caption{Details on internal Amgen datasets. Note: ADME stands for absorption, distribution, metabolism, and excretion.}
\label{tab:amgen_datasets}
\end{table}

\begin{figure}
\centering
\begin{subfigure}{\textwidth}
  \centering
  \includegraphics[width=\linewidth]{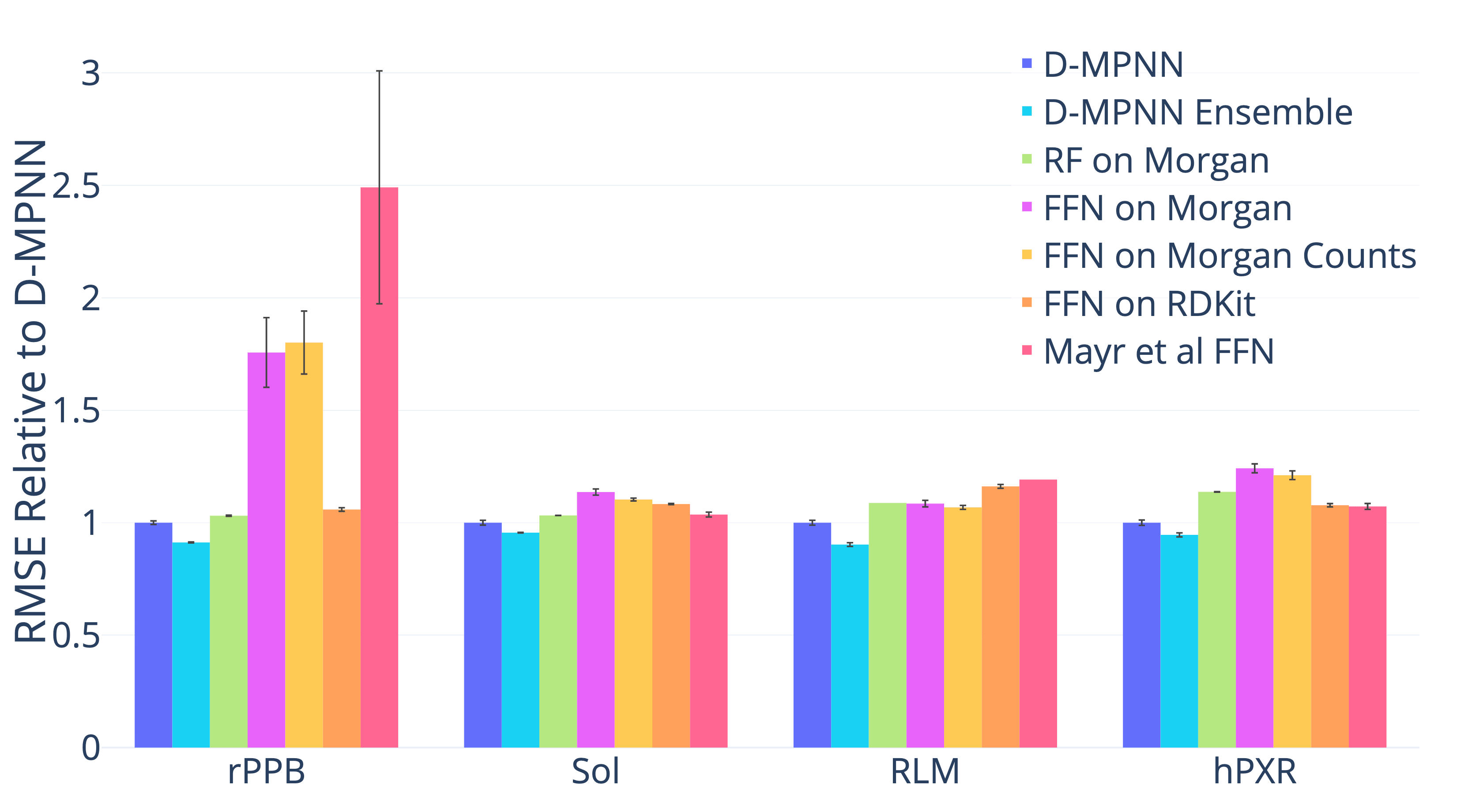}
  \caption{Regression Datasets (lower = better).}
  \label{fig:amgen_baselinesa}
\end{subfigure}%
\vskip\baselineskip
\begin{subfigure}{\textwidth}
  \centering
  \includegraphics[width=\linewidth]{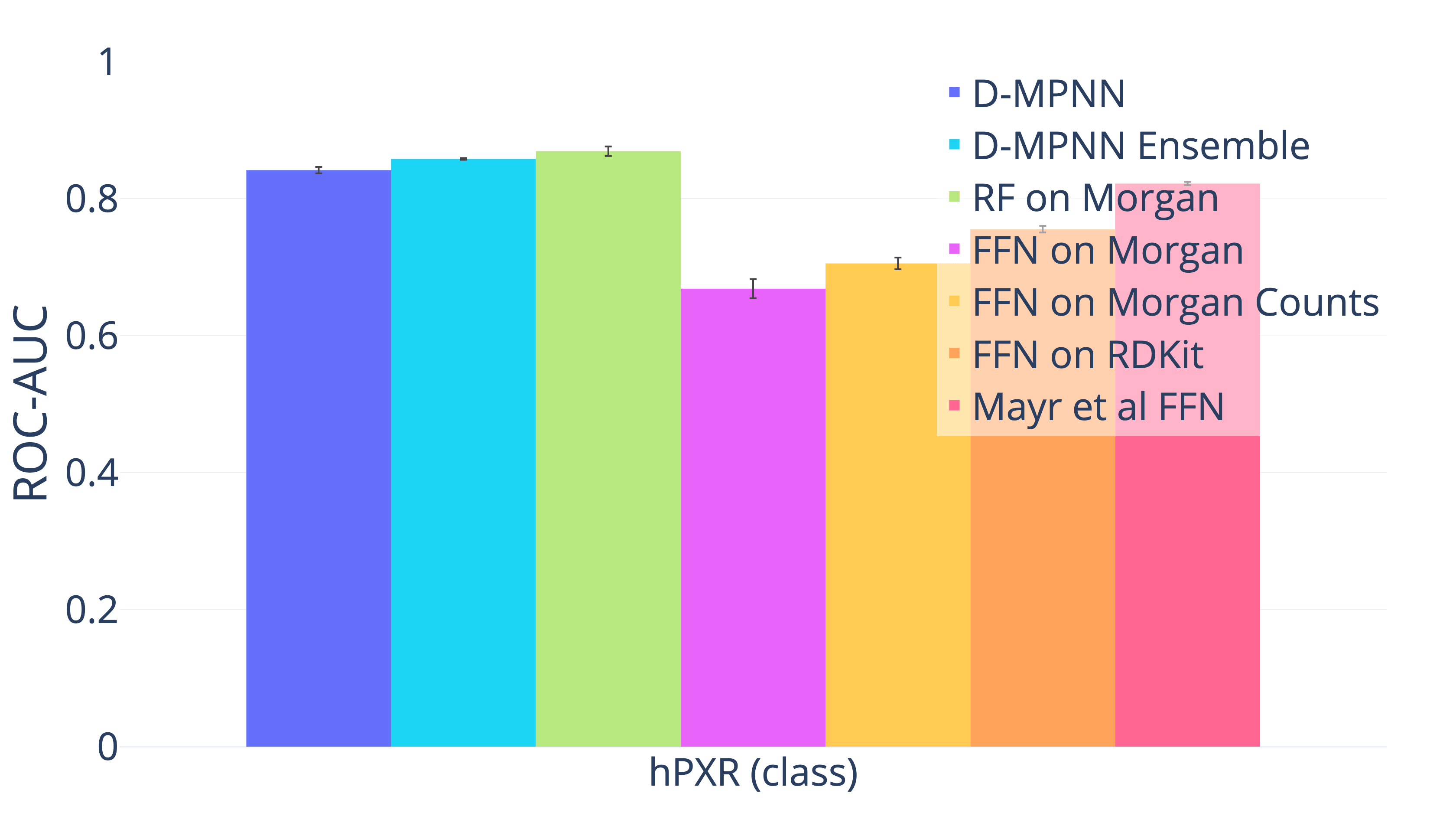}
  \caption{Classification Datasets (higher = better).}
  \label{fig:amgen_baselinesb}
\end{subfigure}
\caption{Comparison of our D-MPNN against baseline models on Amgen internal datasets on a chronological data split. D-MPNN outperforms almost all of the baselines. Note that the ensembles were ensembles of 3 models rather than 5 for the Amgen datasets only. Also note that RF on Morgan and Mayr et al FFN were only run once on RLM.}
\label{fig:amgen_baselines}
\end{figure}

For each dataset, we evaluate on a chronological split. Our model outperforms the baselines on 4 out of the 5 datasets, as shown in Figure \ref{fig:amgen_baselines}. Thus our D-MPNN's strong performance on scaffold splits of public datasets can translate well to chronological splits of private industry datasets.

\FloatBarrier

\subsubsection{BASF}

We ran our model on 10 highly related quantum mechanical datasets from BASF. Each dataset contains 13 properties calculated on the same 30,733 molecules, varying the solvent in each dataset. Dataset details are in Table \ref{tab:basf_datasets}.

\FloatBarrier

\begin{table}[]
\resizebox{\textwidth}{!}{
\centering
\begin{tabular}{|c|c|c|c|c|c|}
\hline
\textbf{Category}          & \textbf{Dataset}       & \textbf{Tasks} & \textbf{Task Type} & \textbf{\# Compounds} & \textbf{Metric} \\
\Xhline{2\arrayrulewidth}
Quantum Mechanics & Benzene         & 13    & regression   & 30,733        & R\textsuperscript{2}     \\
Quantum Mechanics & Cyclohexane     & 13    & regression   & 30,733        & R\textsuperscript{2}     \\
Quantum Mechanics & Dichloromethane & 13    & regression   & 30,733        & R\textsuperscript{2}     \\
Quantum Mechanics & DMSO            & 13    & regression   & 30,733        & R\textsuperscript{2}     \\
Quantum Mechanics & Ethanol         & 13    & regression   & 30,733        & R\textsuperscript{2}     \\
Quantum Mechanics & Ethyl acetate   & 13    & regression   & 30,733        & R\textsuperscript{2}     \\
Quantum Mechanics & H2O             & 13    & regression   & 30,733        & R\textsuperscript{2}     \\
Quantum Mechanics & Octanol         & 13    & regression   & 30,733        & R\textsuperscript{2}     \\
Quantum Mechanics & Tetrahydrofuran & 13    & regression   & 30,733        & R\textsuperscript{2}     \\
Quantum Mechanics & Toluene         & 13    & regression   & 30,733        & R\textsuperscript{2}     \\
\hline
\end{tabular}}
\caption{Details on internal BASF datasets. Note: R\textsuperscript{2} is the square of Pearson's correlation coefficient.}
\label{tab:basf_datasets}
\end{table}

\begin{figure}
    \centering
    \includegraphics[width=\linewidth]{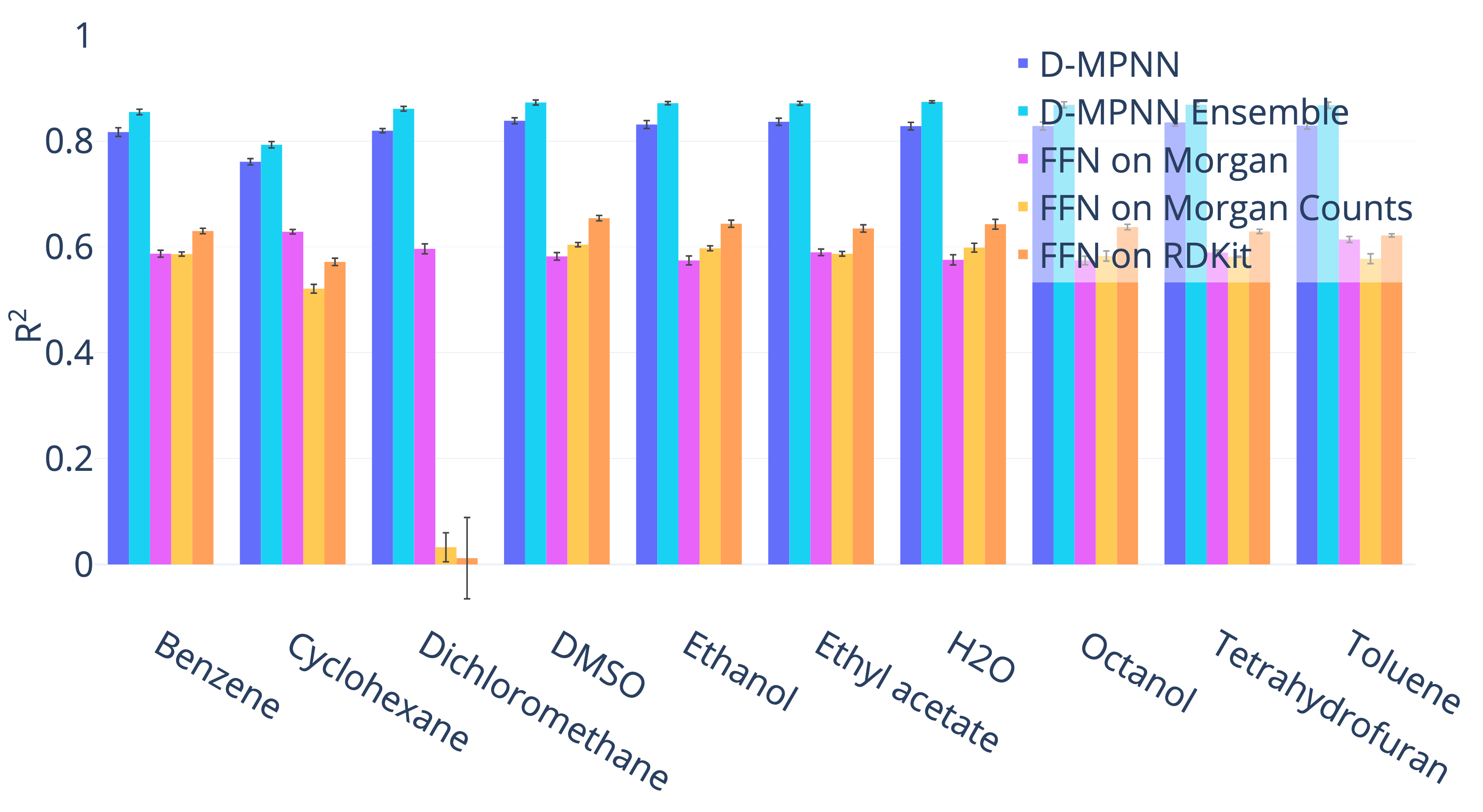}
    \caption{Comparison of our D-MPNN against baseline models on BASF internal regression datasets on a scaffold data split (higher = better). Our D-MPNN outperforms all baselines.}
    \label{fig:basf_baselines}
\end{figure}

For these datasets, we used a scaffold-based split because a chronological split was unavailable. We found that the model of \citeauthor{mayr2018chembl}\cite{mayr2018chembl} is numerically unstable on these datasets, and we therefore omit it from the comparison below. Once again we find that our model, originally designed to succeed on a wide range of public datasets, is robust enough to transfer to proprietary datasets as shown in Figure \ref{fig:basf_baselines}.

\FloatBarrier

\subsubsection{Novartis}

\begin{figure}
    \centering
    \includegraphics[width=\linewidth]{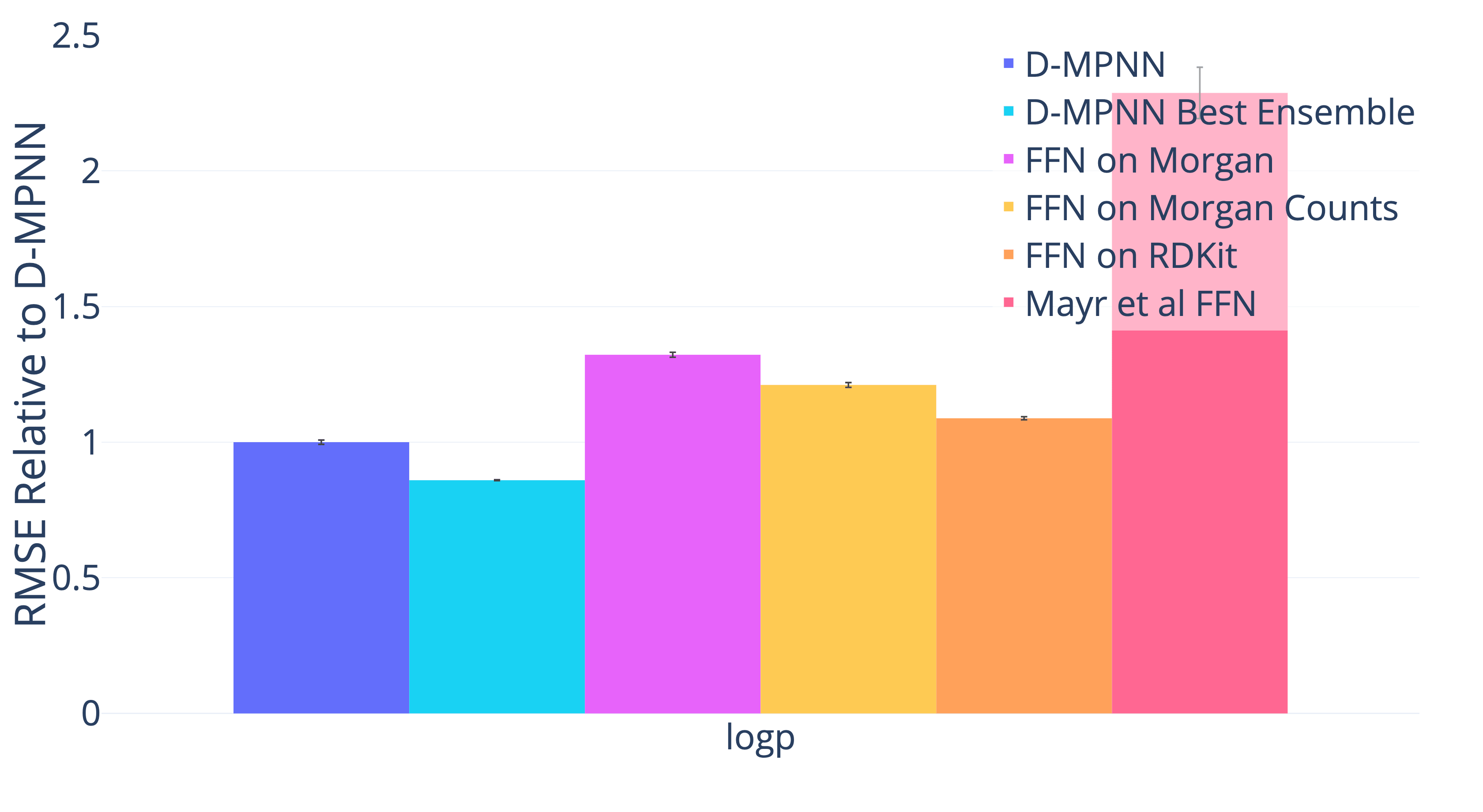}
    \caption{Comparison of our D-MPNN against baseline models on the Novartis internal regression dataset on a chronological data split (lower = better). Our D-MPNN outperforms all baseline models.}
    \label{fig:novartis_baselines}
\end{figure}

Finally, we ran our model on one proprietary dataset from Novartis as described in Table \ref{tab:novartis_datasets}. As with other proprietary datasets, our D-MPNN outperforms the other baselines as shown in Figure \ref{fig:novartis_baselines}.

\FloatBarrier
\begin{table}[]
\centering
\begin{tabular}{|c|c|c|c|c|c|}
\hline
\textbf{Category}           & \textbf{Dataset}   & \textbf{Tasks} & \textbf{Task Type} & \textbf{\# Compounds} & \textbf{Metric} \\
\Xhline{2\arrayrulewidth}
Physical Chemistry & logP      & 1 & regression   & 20,294         & RMSE   \\
\hline
\end{tabular}
\caption{Details on the internal Novartis dataset.}
\label{tab:novartis_datasets}
\end{table}

\subsection{Experimental Error}

\begin{figure}
  \centering
  \includegraphics[width=\linewidth]{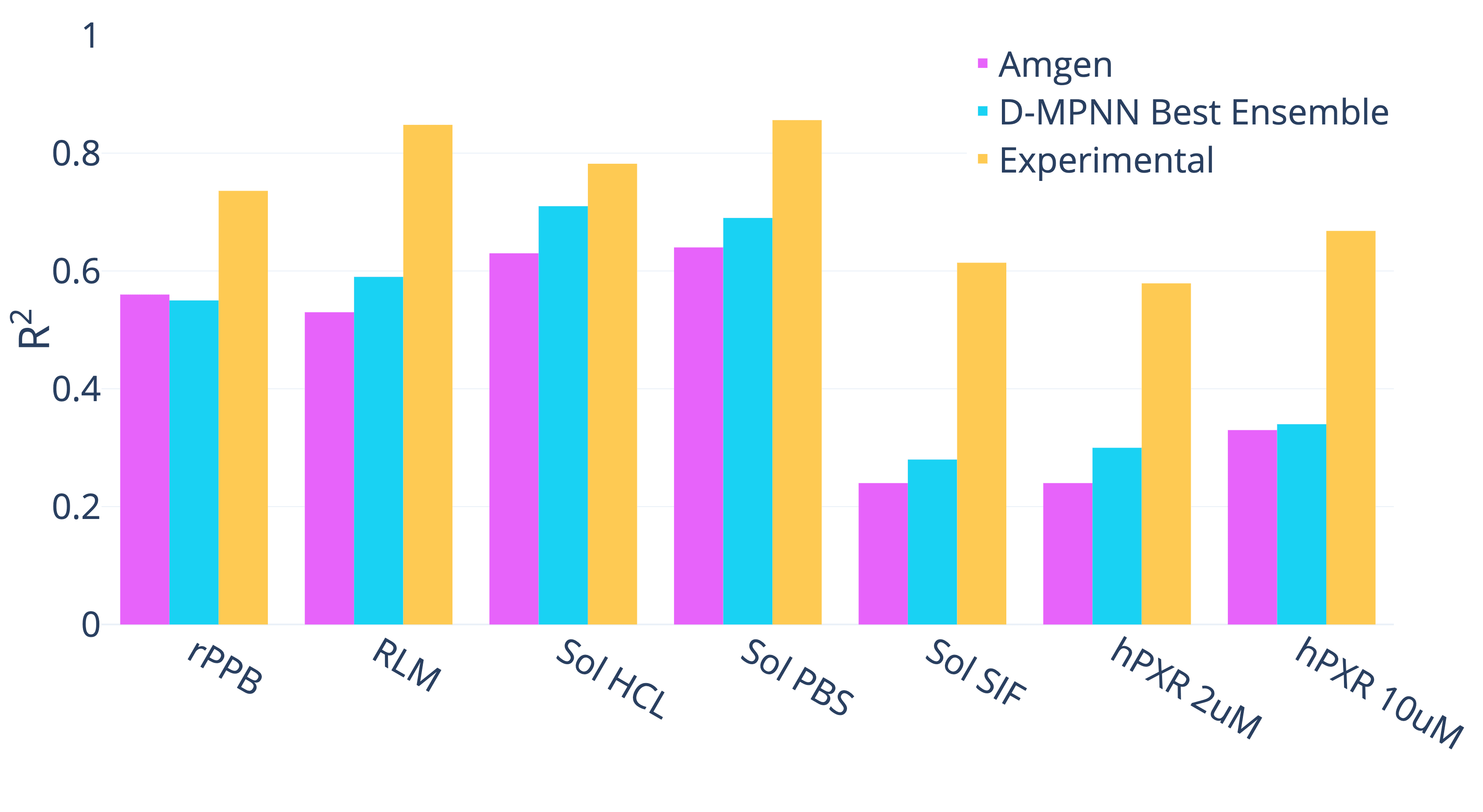}
  \caption{Comparison of Amgen's internal model and our D-MPNN (evaluated using a single run on a chronological split) to experimental error (higher = better). Note that the experimental error is not evaluated on the exact same time split as the two models since it can only be measured on molecules which were tested more than once, but even so the difference in performance is striking.}
  \label{fig:amgen_experimental}
\end{figure}

As a final ``oracle'' baseline, we compare our model's performance with to an experimental upper upper bound: the agreement between multiple runs of the same assay, which we refer to as the \textit{experimental error}. Figure \ref{fig:amgen_experimental} shows the R\textsuperscript{2} of our model on the private Amgen regression datasets together with the experimental error; in addition, this graph shows the performance of Amgen's internal model using expert-crafted descriptors. Both models remain far less accurate than the corresponding ground truth assays. Thus there remains significant space for further performance improvement in the future. 

\FloatBarrier

\subsection{Analysis of Split Type}

We now justify our use of scaffold splits for performance evaluation. The ultimate goal of building a property prediction model is to predict properties on new chemistry in order to aid the search for drugs from new classes of molecules. On proprietary company datasets, performance on new chemistry is evaluated using a chronological split of the data, i.e., everything before a certain date serves as the training set while everything after that date serves as the test set. This approximates model performance on molecules that chemists are likely to investigate in the future. Since chronological data is typically unavailable for public datasets, we investigate whether we can use our scaffold split as a reasonable proxy for a chronological split, following the work of \citeauthor{sheridan2015relative}\cite{sheridan2015relative}. 

\begin{figure}
    \centering
    \includegraphics[width=\linewidth]{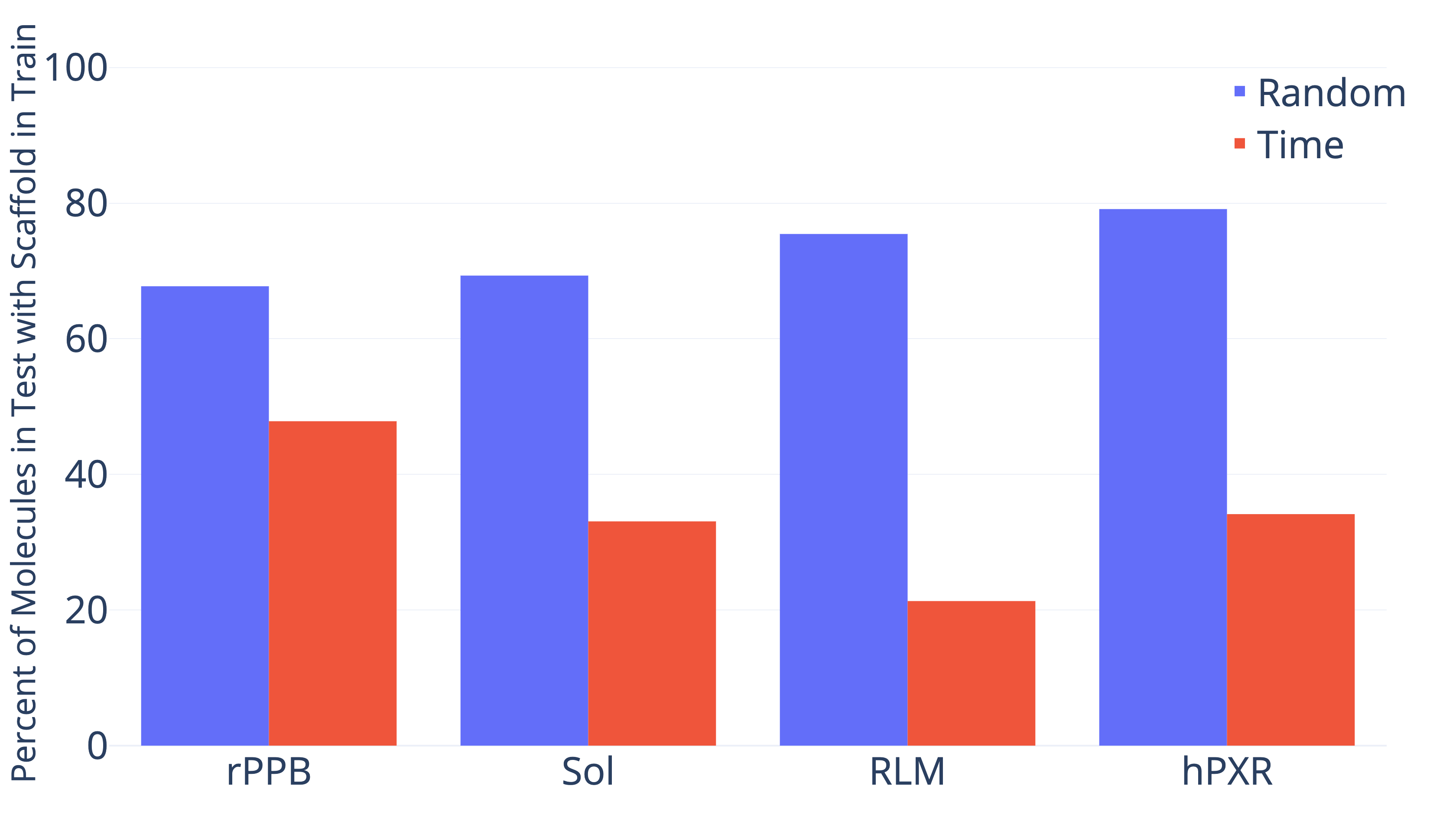}
    \caption{Overlap of molecular scaffolds between the train and test sets for a random or chronological split of four Amgen regression datasets. Overlap is defined as the percent of molecules in the test set which share a scaffold with a molecule in the train set.}
    \label{fig:amgen_scaffold_overlap}
\end{figure}

Figure \ref{fig:amgen_scaffold_overlap} provides motivation for a scaffold split approach. As illustrated in the figure, train and test sets according to a chronological split share fewer molecular scaffolds than train and test sets split randomly. Since our scaffold split enforces zero molecular scaffold overlap between the train and test set, it should ideally provide a split that is at least as difficult as a chronological split.

\begin{figure}
    \centering
    \includegraphics[width=\linewidth]{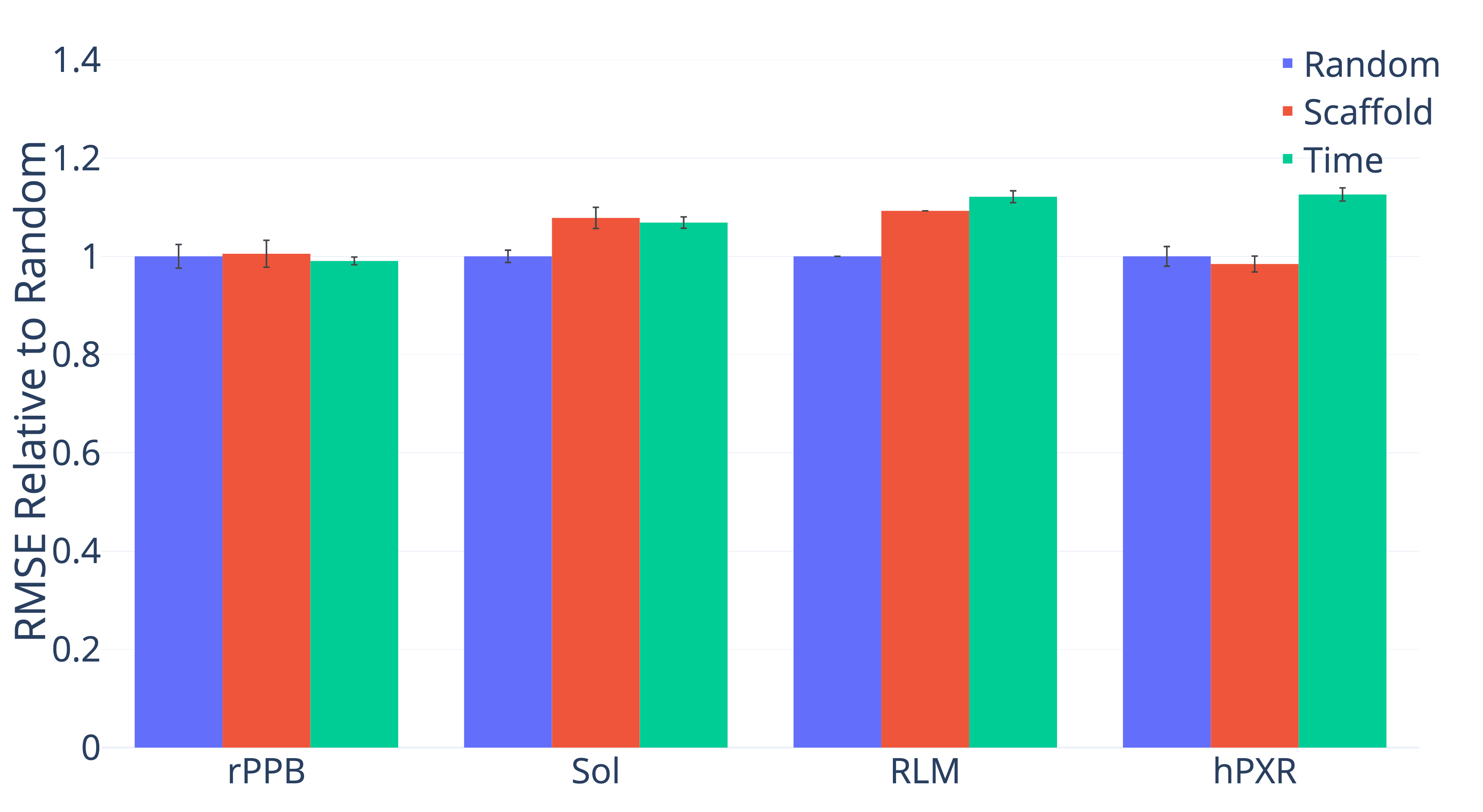}
    \caption{Performance of D-MPNN on four Amgen regression datasets according to three methods of splitting the data (lower = better). The chronological split is significantly harder than both random and scaffold on Sol and hPXR, while the scaffold split is significantly harder than the random split on Sol only.}
    \label{fig:amgen_split_type}
\end{figure}

\begin{figure}
    \centering
    \includegraphics[width=\linewidth]{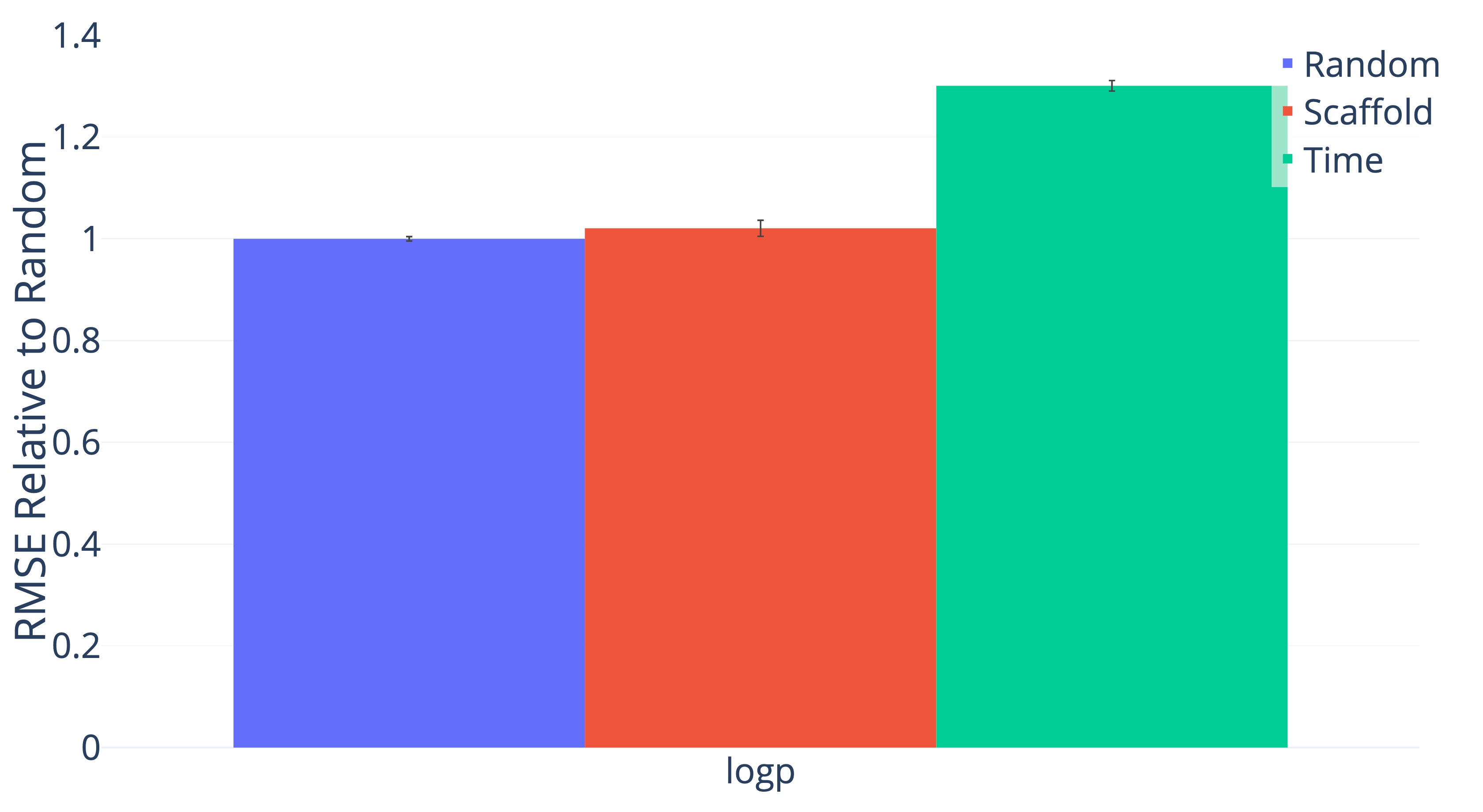}
    \caption{Performance of D-MPNN on the Novartis regression dataset according to three methods of splitting the data (lower = better). The chronological split is significantly harder than the random split while the scaffold split is not.}
    \label{fig:novartis_split_type}
\end{figure}

\begin{figure}
    \centering
    \includegraphics[width=\linewidth]{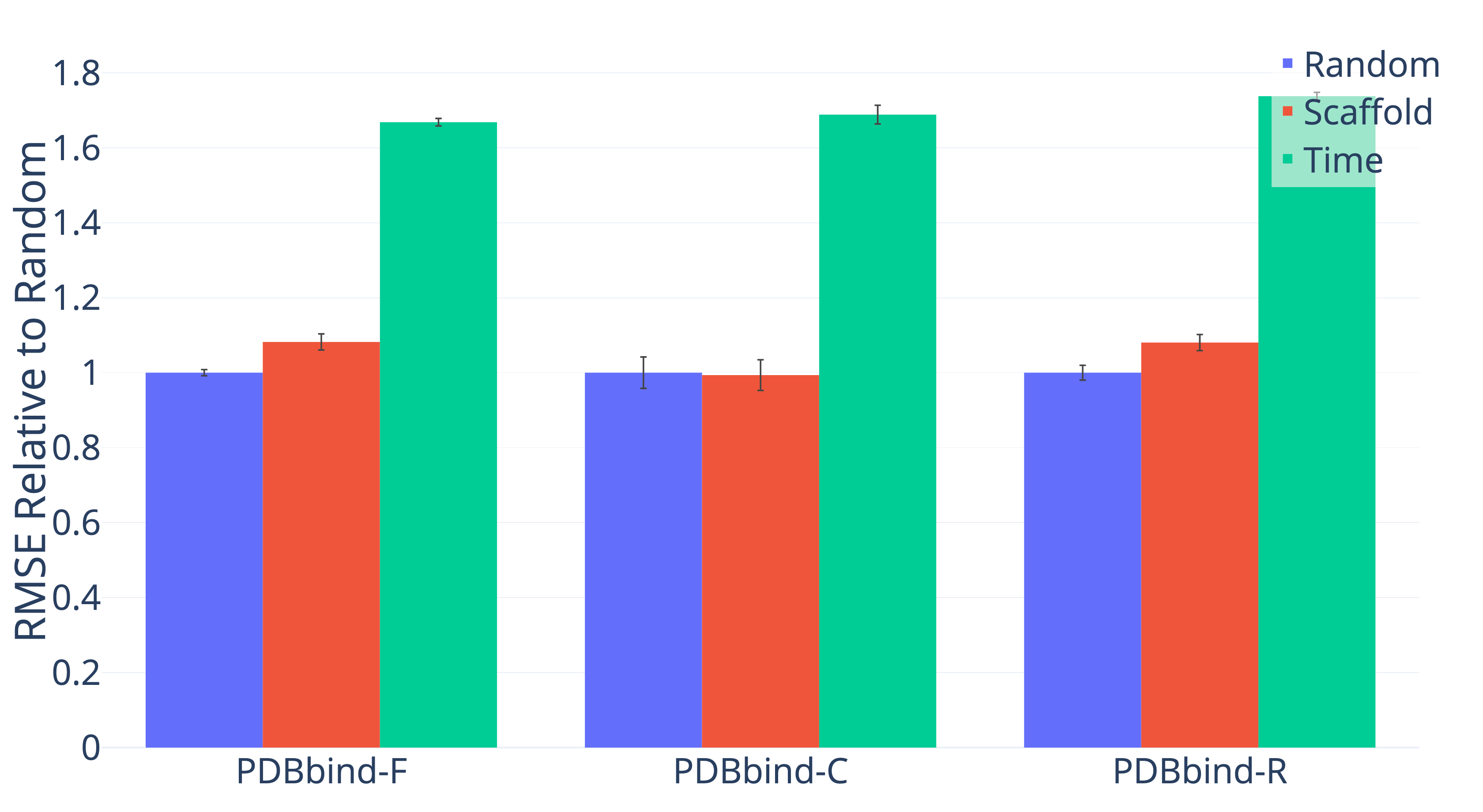}
    \caption{Performance of D-MPNN on the full (F), core (C), and refined (R) subsets of the PDBbind dataset according to three methods of splitting the data (lower = better). The chronological and scaffold splits are significantly harder than the random split in all cases except for the PDBbind-C scaffold split.}
    \label{fig:pdbbind_split_type}
\end{figure}

As illustrated in Figures \ref{fig:amgen_split_type}, \ref{fig:novartis_split_type}, and \ref{fig:pdbbind_split_type}, performance on our scaffold split is on average closer to performance on a chronological split on proprietary datasets from Amgen and Novartis and on the public PDBbind datasets. However, the results are noisy due to the nature of chronological splitting, where we only have a single data split, as opposed to random and scaffold splitting, which both have a random component and can generate different splits depending on the random seed. We can alleviate the problem with noise in chronological datasets by using a sliding time window to get different equally-sized splits, at the cost of significantly decreasing the dataset size. We report results on such sliding window splits in the Supporting Information, as the conclusions from these splits are qualitatively similar to those in the main paper.

\begin{figure}
\centering
\begin{subfigure}{\textwidth}
  \centering
  \includegraphics[width=\linewidth]{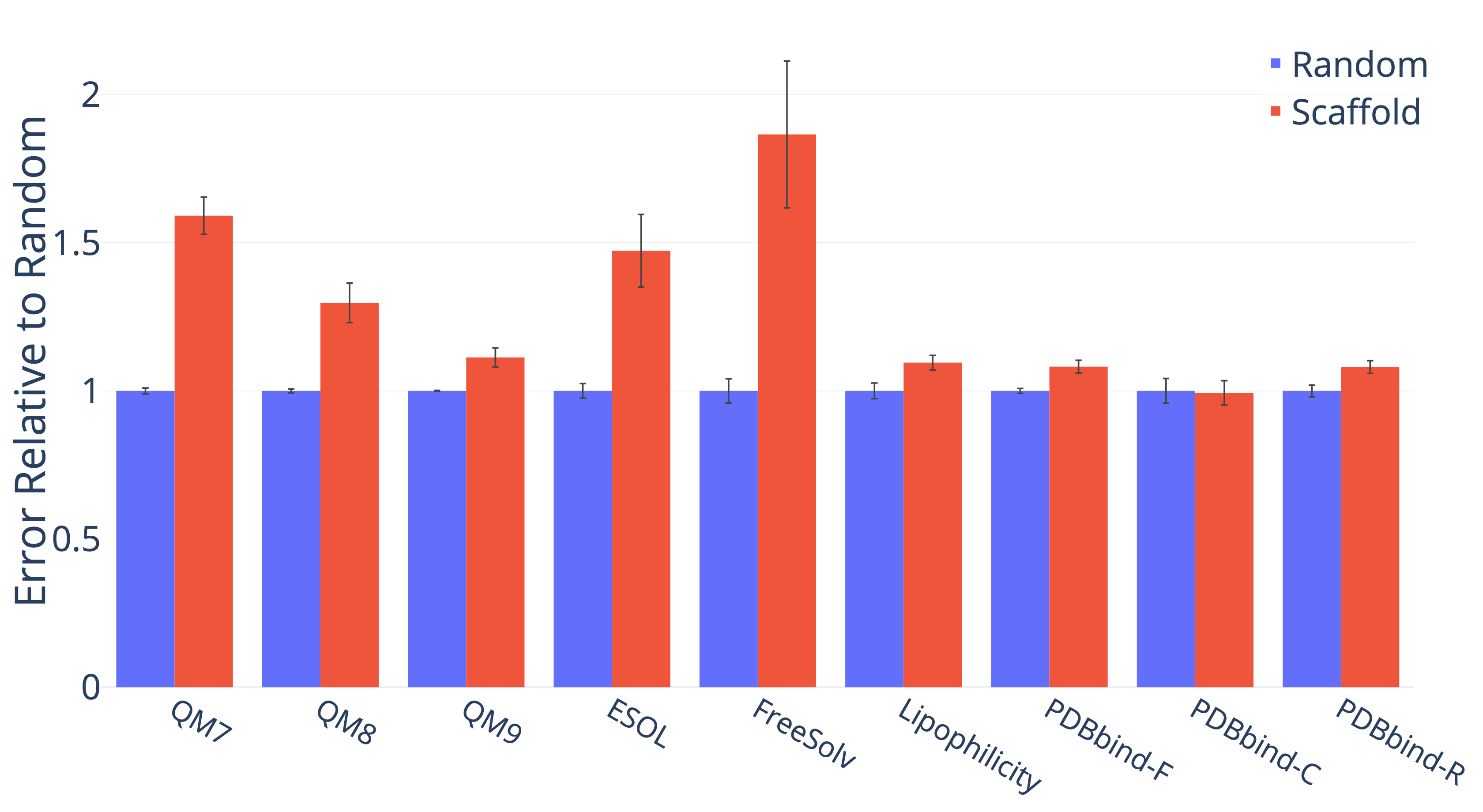}
  \caption{Regression Datasets (lower = better).}
  \label{fig:split_typea}
\end{subfigure}%
\vskip\baselineskip
\begin{subfigure}{\textwidth}
  \centering
  \includegraphics[width=\linewidth]{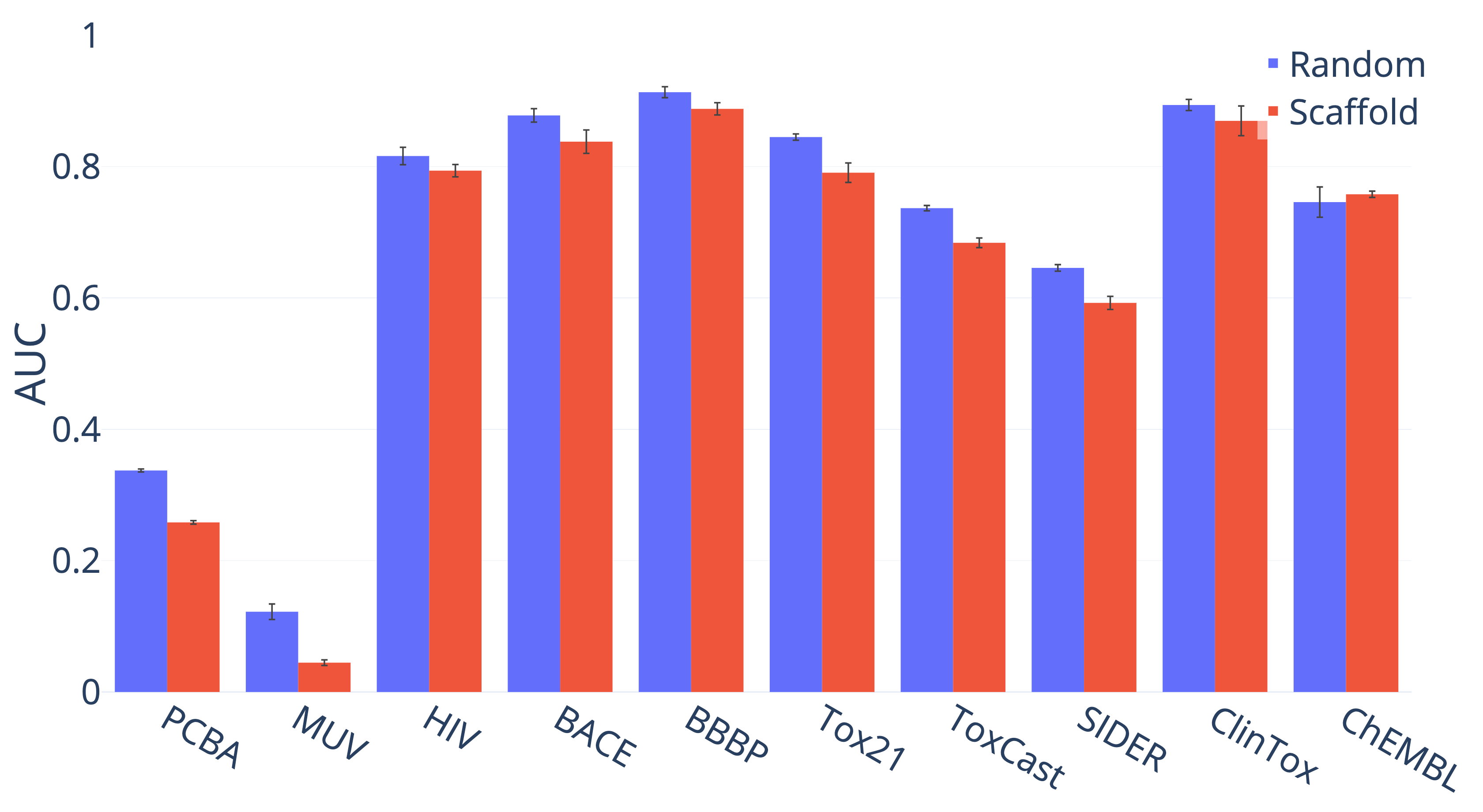}
  \caption{Classification Datasets (higher = better).}
  \label{fig:split_typeb}
\end{subfigure}
\caption{Performance of D-MPNN on random and scaffold splits for several public datasets.  Only the results on PDBbind-C, HIV, ClinTox, and ChEMBL are not statistically significant.}
\label{fig:split_type}
\end{figure}

Figure \ref{fig:split_type} shows the difference between a random split and a scaffold split on the publicly available datasets, further demonstrating that a scaffold split generally results in a more difficult, and ideally more useful, measure of performance. Therefore, all of our results are reported on a scaffold split rather than a random split in order to better reflect the generalization ability of our model on new chemistry. Nevertheless, it should be emphasized that the chronological split is still the ideal split on which to evaluate when it is available. Thus we additionally report the results on chronological splits on all datasets where they are available. 

Overall, our results confirm the findings of \citeauthor{sheridan2015relative}\cite{sheridan2015relative} that scaffold and chronological splits are more difficult than random splits, and hence scaffold splits should be preferred over random splits during evaluation. Our findings differ somewhat from those in \citeauthor{sheridan2015relative}\cite{sheridan2015relative} in that we find some evidence that chronological splits may actually be harder than scaffold splits. However, owing to the small number of datasets where chronological splits are available, further investigation is necessary on this point, ideally on a larger range of datasets.

\FloatBarrier

% CWC: I would try to clean up these plots if possible. Uniform font choice, size, etc. It is also very uncommon to have titles in figures - the title should be reflected by the caption.

% CWC: same plot formatting suggestion as above. Left plot should be labeled "a" and right "b" with titles in the captions. 

% CWC: is that the right conclusion from these plots though? rPPB-scaffold looks too conservative, and hPXR-scaffold is even more optimistic than random 
% CWC: I know a large part of the evaluation is emphasizing the scaffold split as a proxy for a time split. This should certainly be true for physical property data when you're thinking about exploring a new area of chemical space. For performance on specific biological assays, there is the complication that multiple targets can have different mechanisms of action, and so you wouldn't expect a model trained to recognize performance through mechanism A would generalize to molecules that operate through mechanism B. A small statement about this might be good for the introduction 

\subsection{Ablations}

Finally, we analyze and justify our modeling choices and optimizations. 

\subsubsection{Message Type}

\begin{figure}
\centering
\begin{subfigure}{\textwidth}
  \centering
  \includegraphics[width=\linewidth]{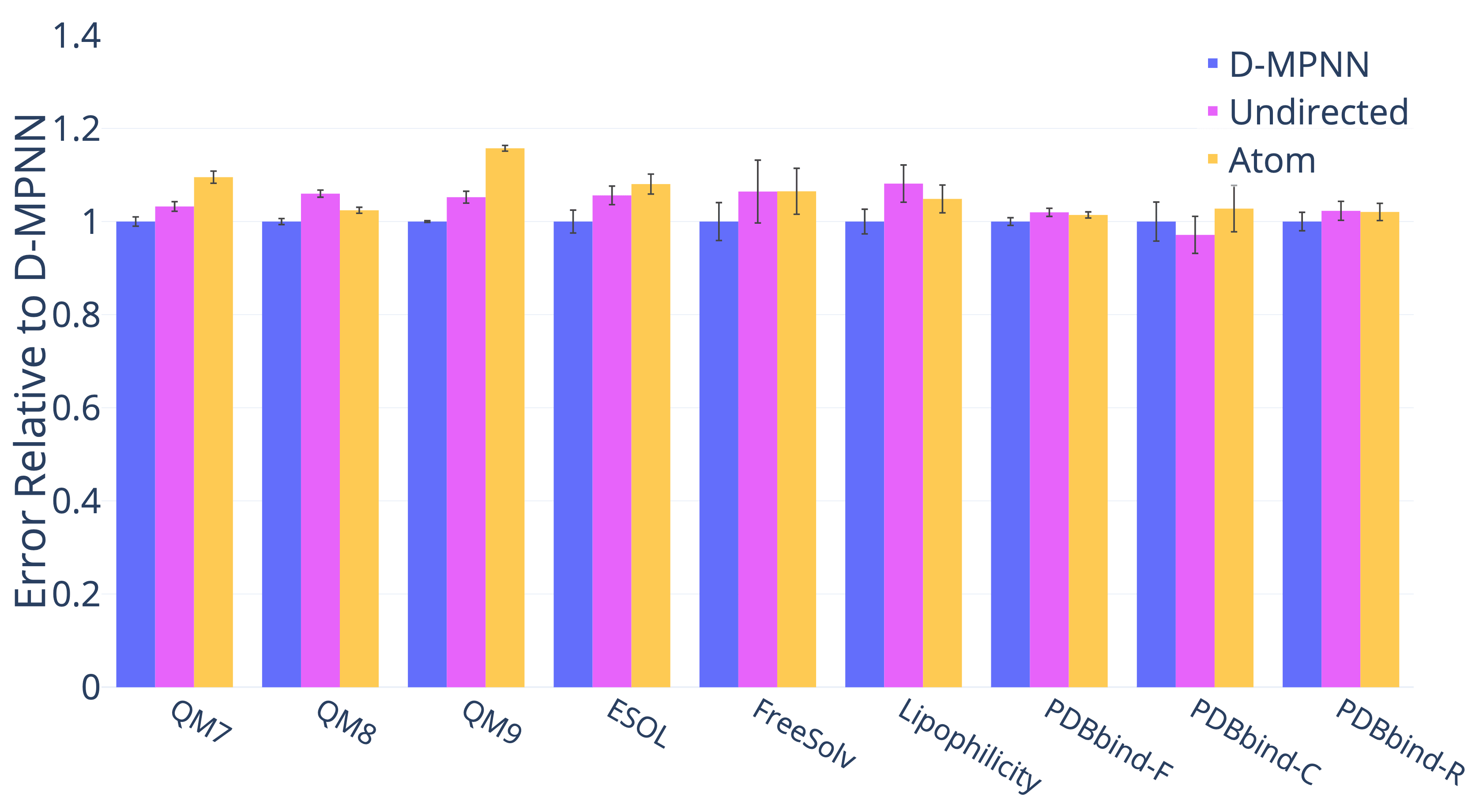}
  \caption{Regression Datasets (lower = better).}
  \label{fig:message_typea}
\end{subfigure}%
\vskip\baselineskip
\begin{subfigure}{\textwidth}
  \centering
  \includegraphics[width=\linewidth]{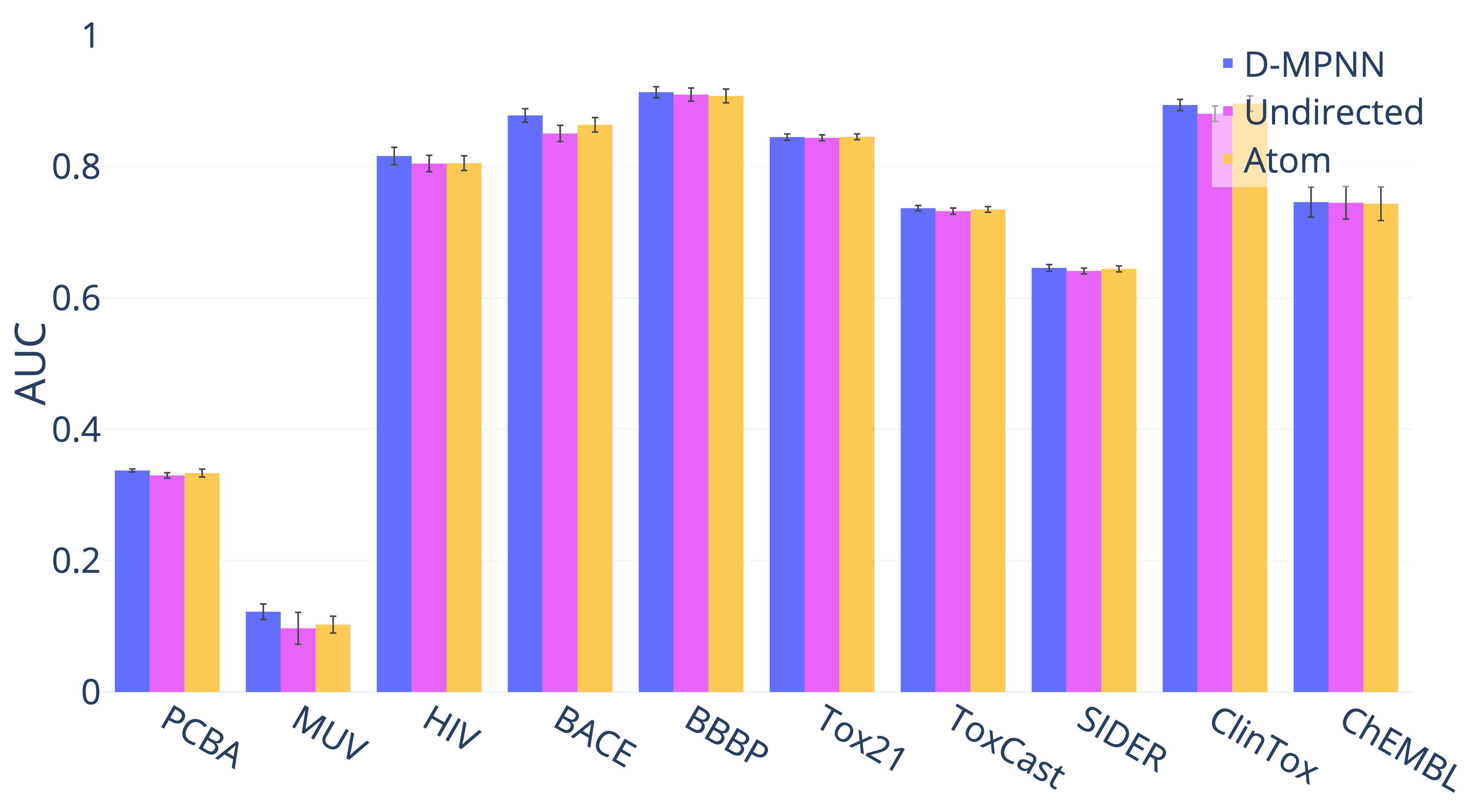}
  \caption{Classification Datasets (higher = better).}
  \label{fig:message_typeb}
\end{subfigure}
\caption{Comparison of performance of different message passing paradigms.}
\label{fig:message_type}
\end{figure}

The most important distinction between our D-MPNN and related work is the nature of the messages being passed across the molecule. Most prior work uses messages centered on atoms whereas our D-MPNN uses messages centered on directed bonds. To isolate the effect of the message passing paradigm on property prediction performance, we implemented message passing on undirected bonds and on atoms as well, as detailed in the Supporting Information and in our code. Figure \ref{fig:message_type} illustrates the differences in performance between these three types of message passing. While on average the method using directed bonds outperforms the alternatives, the results are largely not statistically significant, so more investigation is warranted on this point. 
% CWC: do you have a high-level description of why you chose to try the directed bond approach? the performance can speak for itself, but it might be nice to motivate why this architecture could make sense for chemical problems in 1-2 sentences. 
% KY: added more motivation at beginning of Methods section

\FloatBarrier

\subsubsection{RDKit Features}

\begin{figure}
\centering
\begin{subfigure}{\textwidth}
  \centering
  \includegraphics[width=\linewidth]{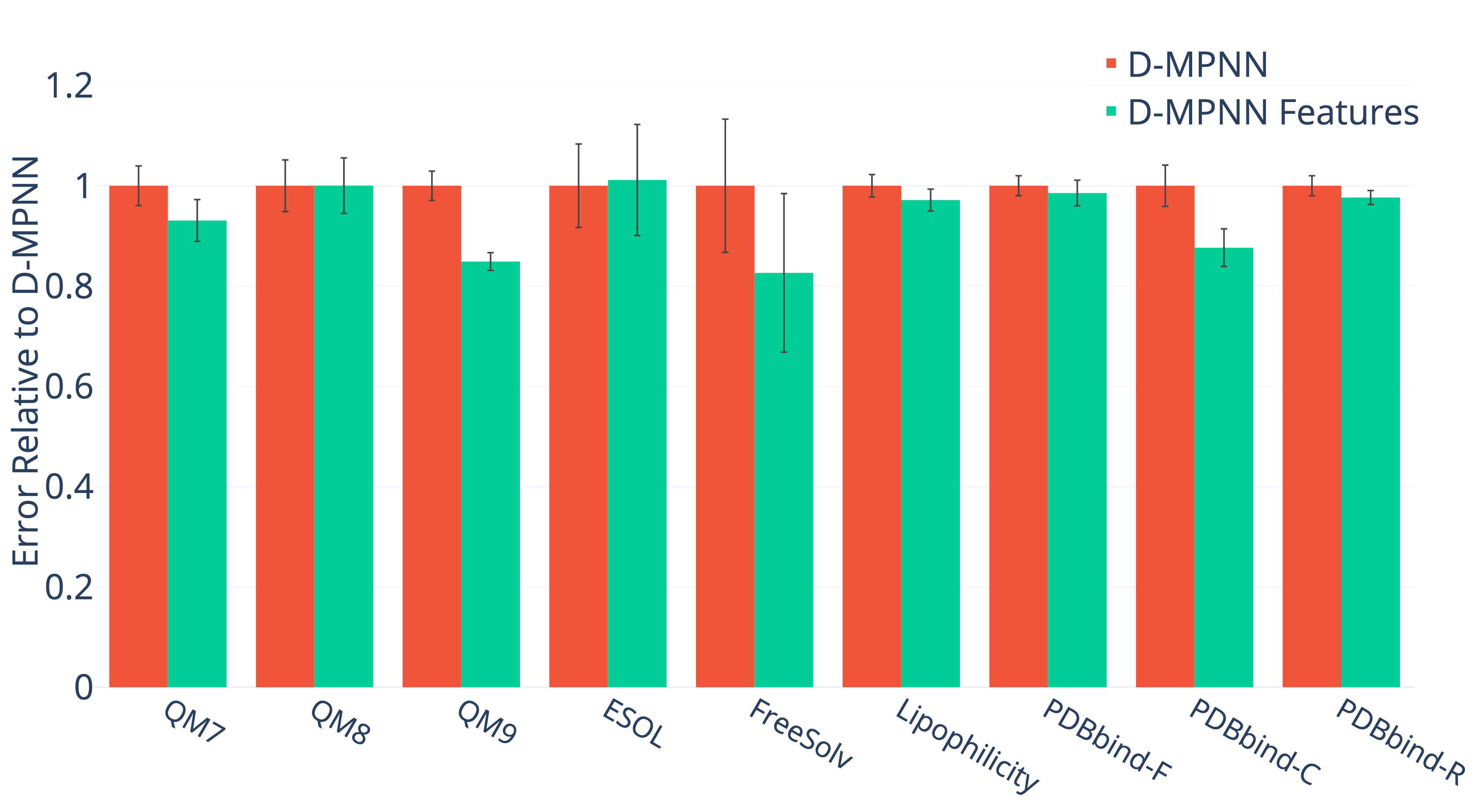}
  \caption{Regression Datasets (lower = better).}
  \label{fig:featuresa}
\end{subfigure}%
\vskip\baselineskip
\begin{subfigure}{\textwidth}
  \centering
  \includegraphics[width=\linewidth]{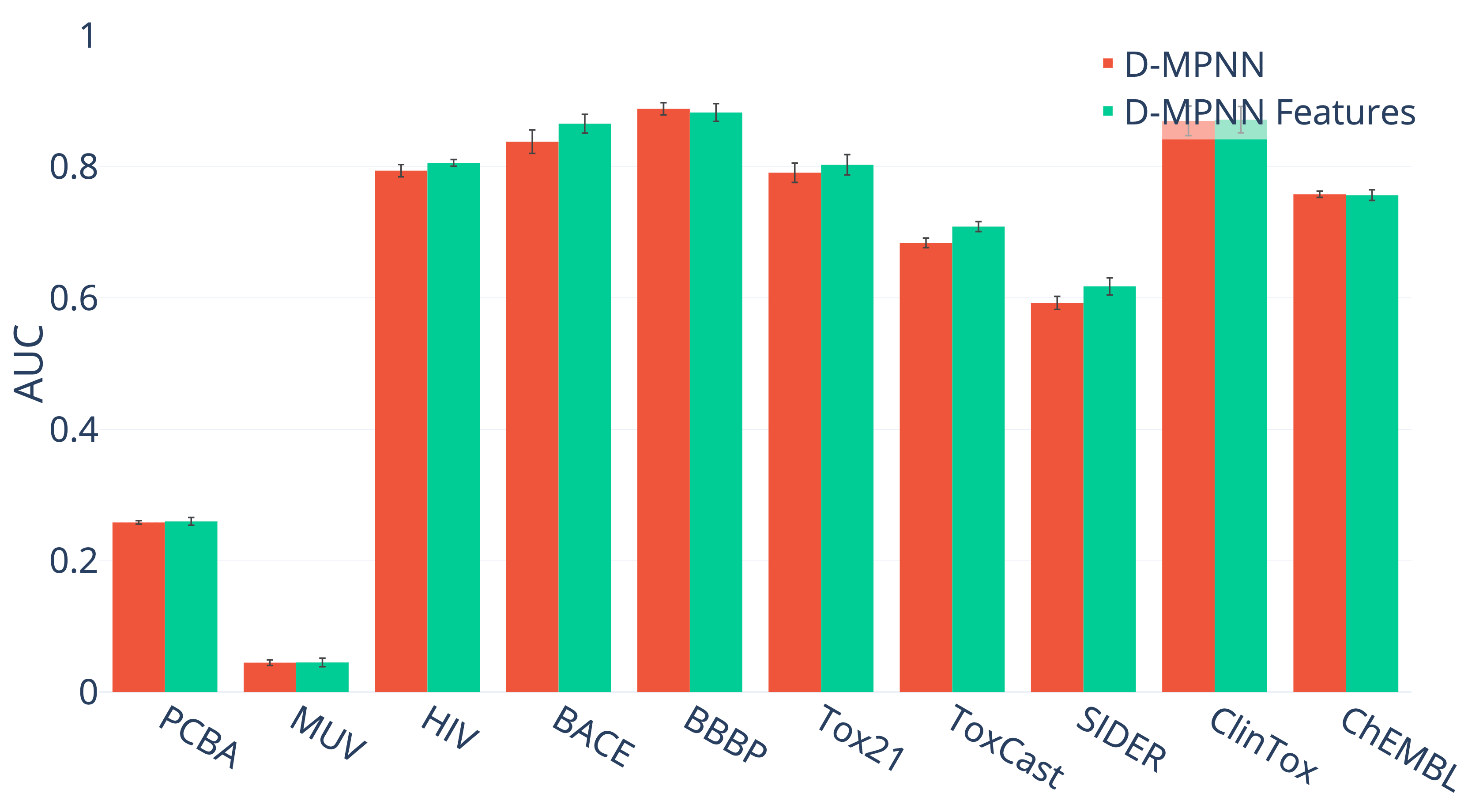}
  \caption{Classification Datasets (higher = better).}
  \label{fig:featuresb}
\end{subfigure}
\caption{Effect of adding molecule-level features generated with RDKit to our model.}
\label{fig:features}
\end{figure}

Next, we examined the impact of adding additional molecule-level features from RDKit to our model. Figure \ref{fig:features} shows the effect on model performance. The results appear to be highly dataset-dependent. Some datasets, such as QM9 and ESOL, show marked improvement with the addition of features, while other datasets, such as PCBA and HIV, actually show worse performance with the features. We hypothesize that this is because the features are particularly relevant to certain tasks while possibly confusing and distracting the model on other tasks. This implies that our model's performance on a given dataset may be further optimized by selecting different features more relevant to the task of interest.

\FloatBarrier

Another interesting trend is the effect of adding features to the three PDBbind datasets. The features appear to help on all three datasets, but the benefit is much more pronounced on the extremely small PDBbind-C (core) dataset than it is on the larger PDBbind-R (refined) and PDBbind-F (full) datasets. This indicates that the features may help compensate for the lack of training data and thus may be particularly relevant in low-data regimes. In particular, we hypothesize that the features may help to regularize a representation derived from a small dataset: because the features are derived from more general chemical knowledge, they implicitly provide the model some understanding of a larger chemical domain. Thus, it is worthwhile to consider the addition of features both when they are particularly relevant to the task of interest and when the dataset is especially small.

\subsubsection{Hyperparameter Optimization}

\begin{figure}
\centering
\begin{subfigure}{\textwidth}
  \centering
  \includegraphics[width=\linewidth]{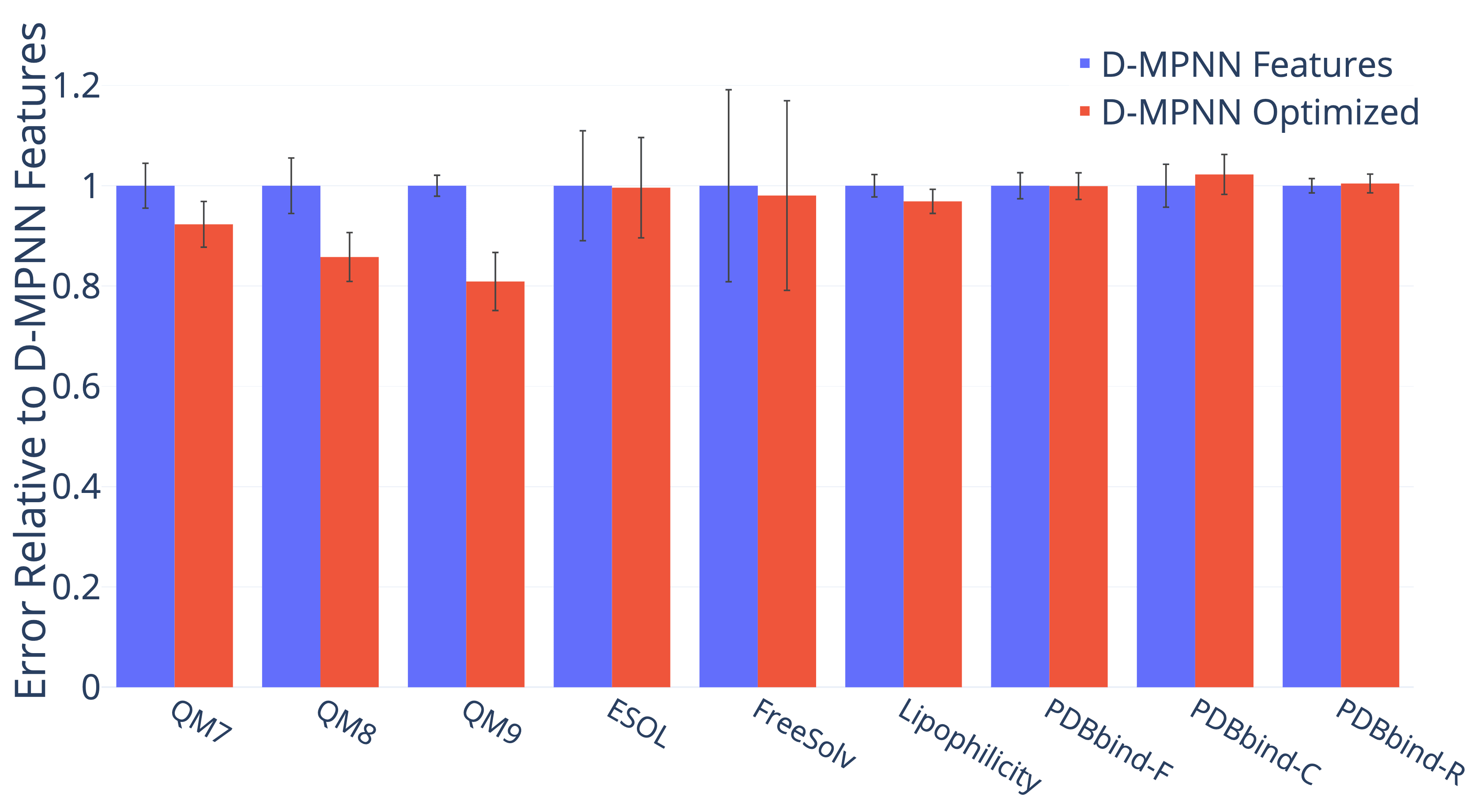}
  \caption{Regression Datasets (lower = better).}
  \label{fig:hyperopta}
\end{subfigure}%
\vskip\baselineskip
\begin{subfigure}{\textwidth}
  \centering
  \includegraphics[width=\linewidth]{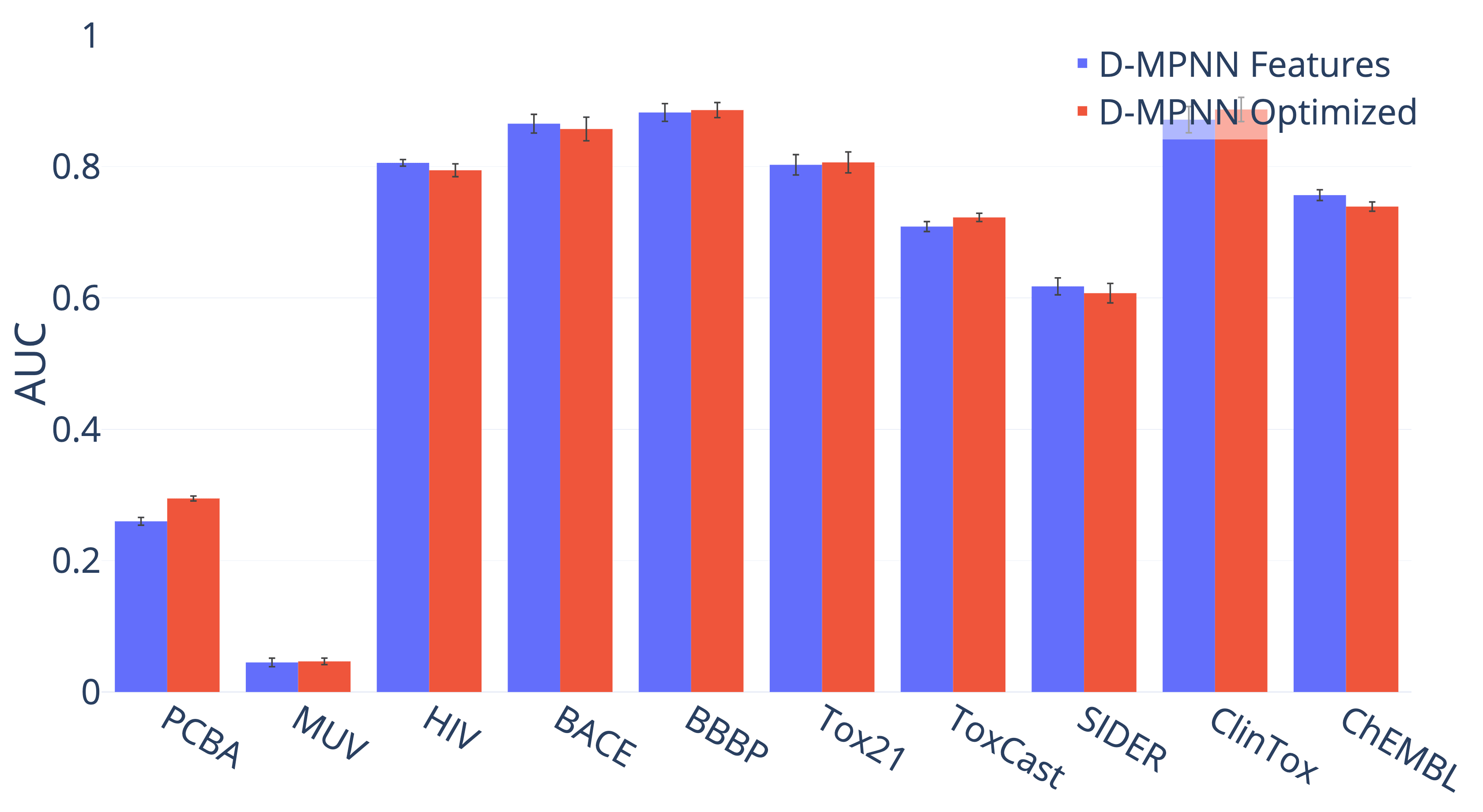}
  \caption{Classification Datasets (higher = better).}
  \label{fig:hyperoptb}
\end{subfigure}
\caption{Effect of performing Bayesian hyperparameter optimization on the depth, hidden size, number of fully connected layers, and dropout of the D-MPNN.}
\label{fig:hyperopt}
\end{figure}

To improve model performance, we performed Bayesian Optimization to select the best model hyperparameters for each dataset. Figure \ref{fig:hyperopt} illustrates the benefit of performing this optimization, as model performance improves on virtually every dataset. Interestingly, some datasets are particularly sensitive to hyperparameters. While most datasets experience a moderate 2-5\% improvement in performance following hyperparameter optimization, the quantum mechanics datasets (QM7, QM8, and QM9) and PCBA see dramatic improvements in performance, with our D-MPNN model performing 37\% better on QM9 after optimization.

\subsubsection{Ensembling}

\begin{figure}
\centering
\begin{subfigure}{\textwidth}
  \centering
  \includegraphics[width=\linewidth]{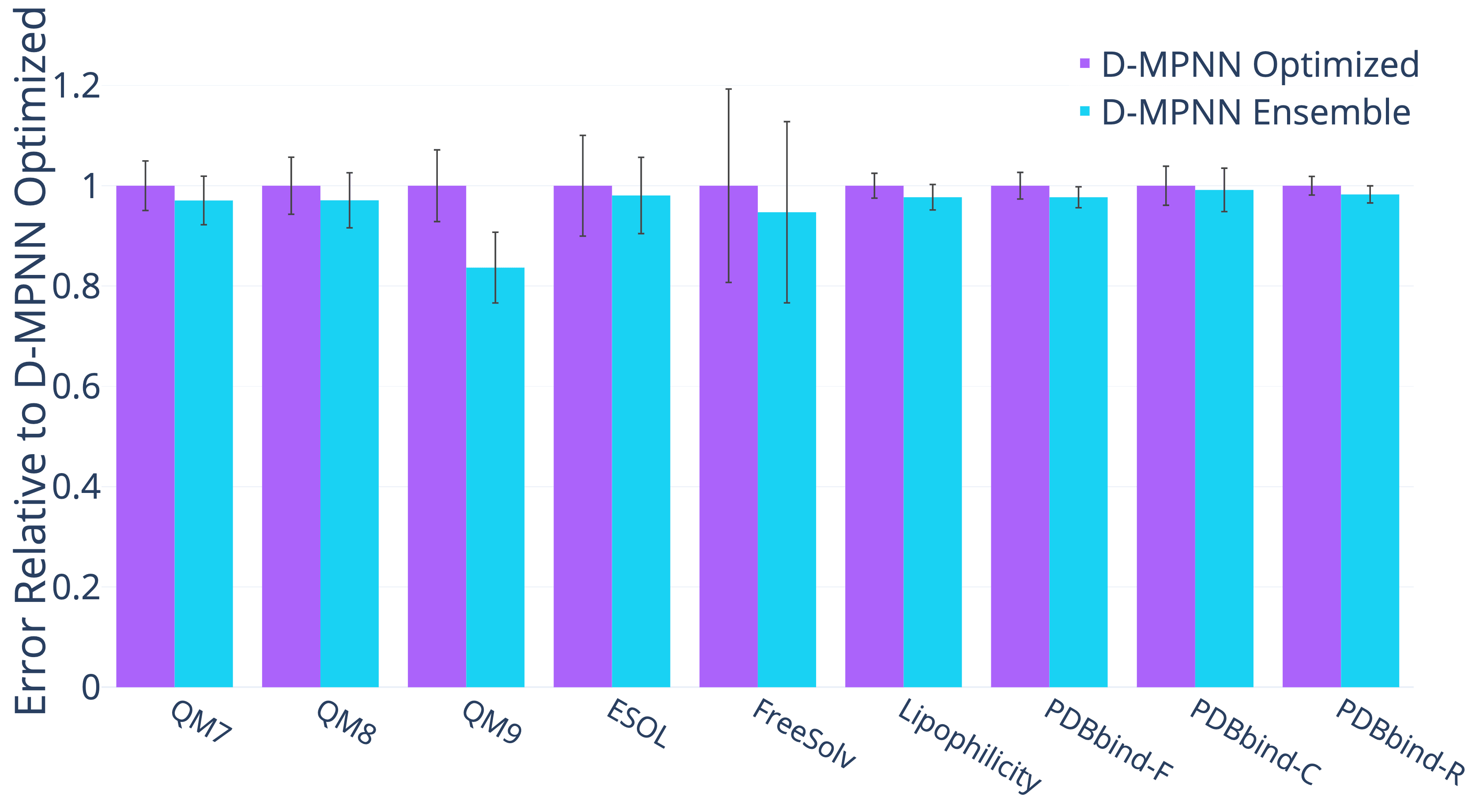}
  \caption{Regression Datasets (lower = better).}
  \label{fig:ensemblea}
\end{subfigure}%
\vskip\baselineskip
\begin{subfigure}{\textwidth}
  \centering
  \includegraphics[width=\linewidth]{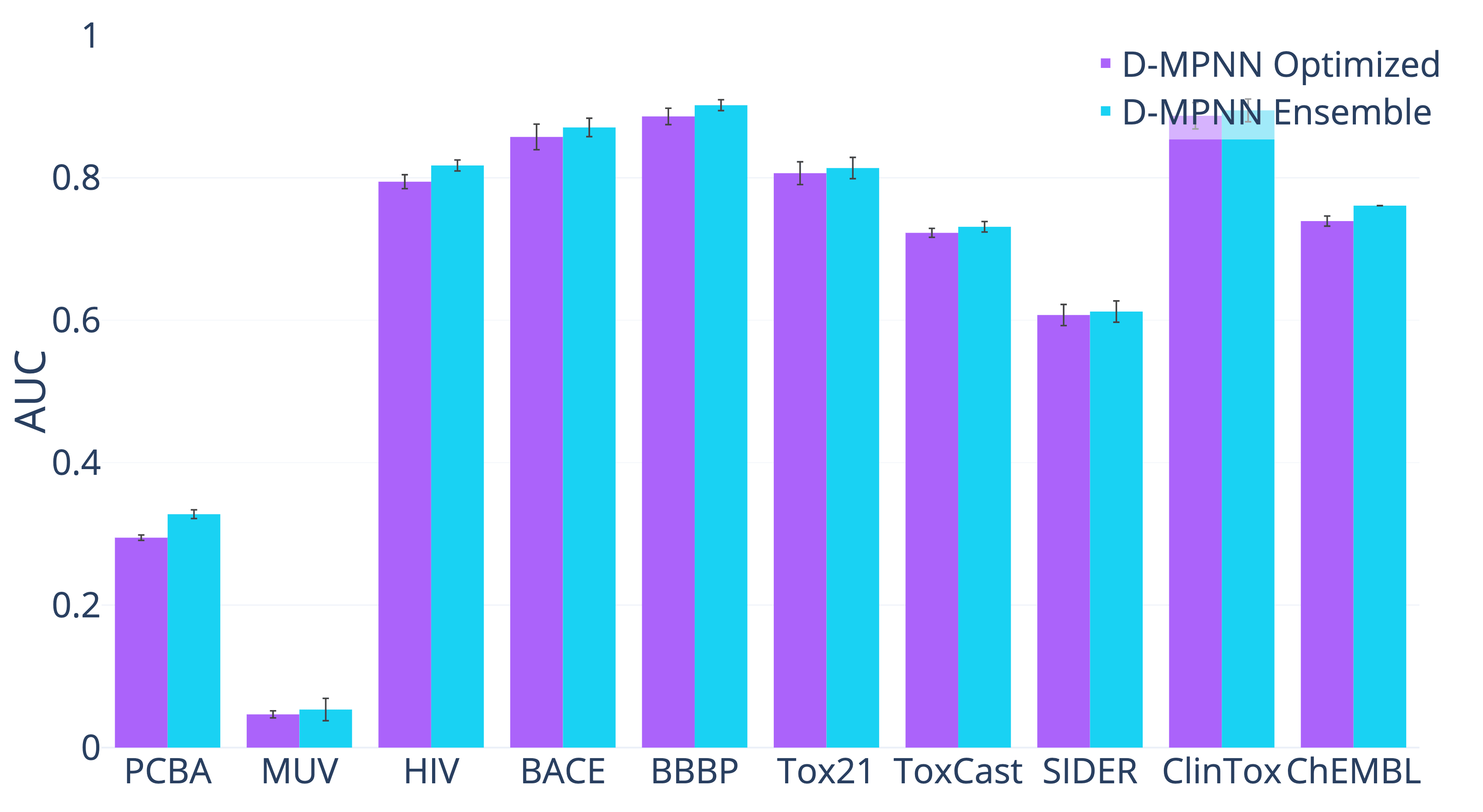}
  \caption{Classification Datasets (higher = better).}
  \label{fig:ensembleb}
\end{subfigure}
\caption{Effect of using an ensemble of five models instead of a single model.}
\label{fig:ensemble}
\end{figure}

To maximize performance, we trained an ensemble of models. For each dataset, we selected the best single model---i.e. the best hyperparameters along with the RDKit features if the features improved performance---and we trained five models instead of one. The results appear in Figure \ref{fig:ensemble}. On most datasets, ensembling only provides a small 1-5\% benefit, but as with hyperparameter optimization, there are certain datasets, particularly the quantum mechanics datasets, which especially benefit from the effect of ensembling.

\FloatBarrier

\begin{figure}
    \centering
    \includegraphics[width=\linewidth]{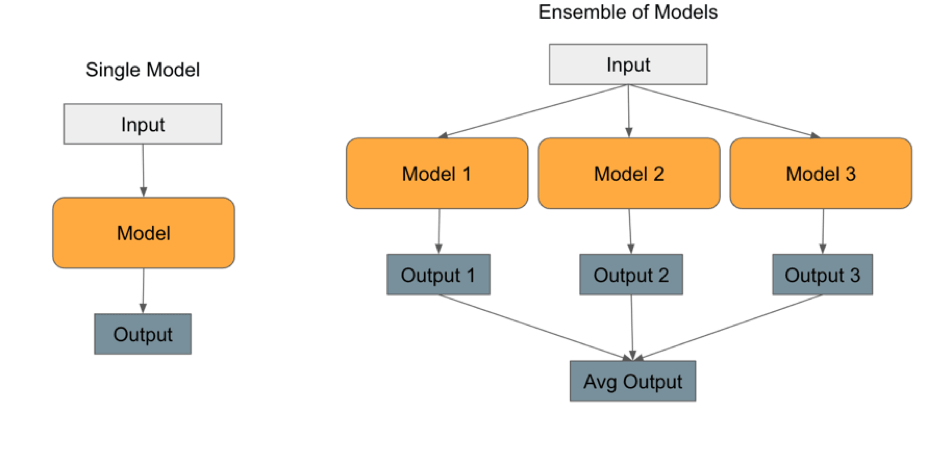}
    \caption{An illustration of ensembling models. On the left is a single model, which takes input and makes a prediction. On the right is an ensemble of 3 models. Each model takes the same input and makes a prediction independently, and then the predictions are averaged to generate the ensemble's prediction.}
    \label{fig:ensemble_diagram}
\end{figure}

While each of the latter three optimizations (RDKit descriptors, hyperparameter optimization, and ensembling) on its own has limited benefits, altogether they significantly improve the model's performance on every dataset except MUV.

\subsubsection{Effect of Data Size}

\begin{figure}
    \centering
    \includegraphics[width=\linewidth]{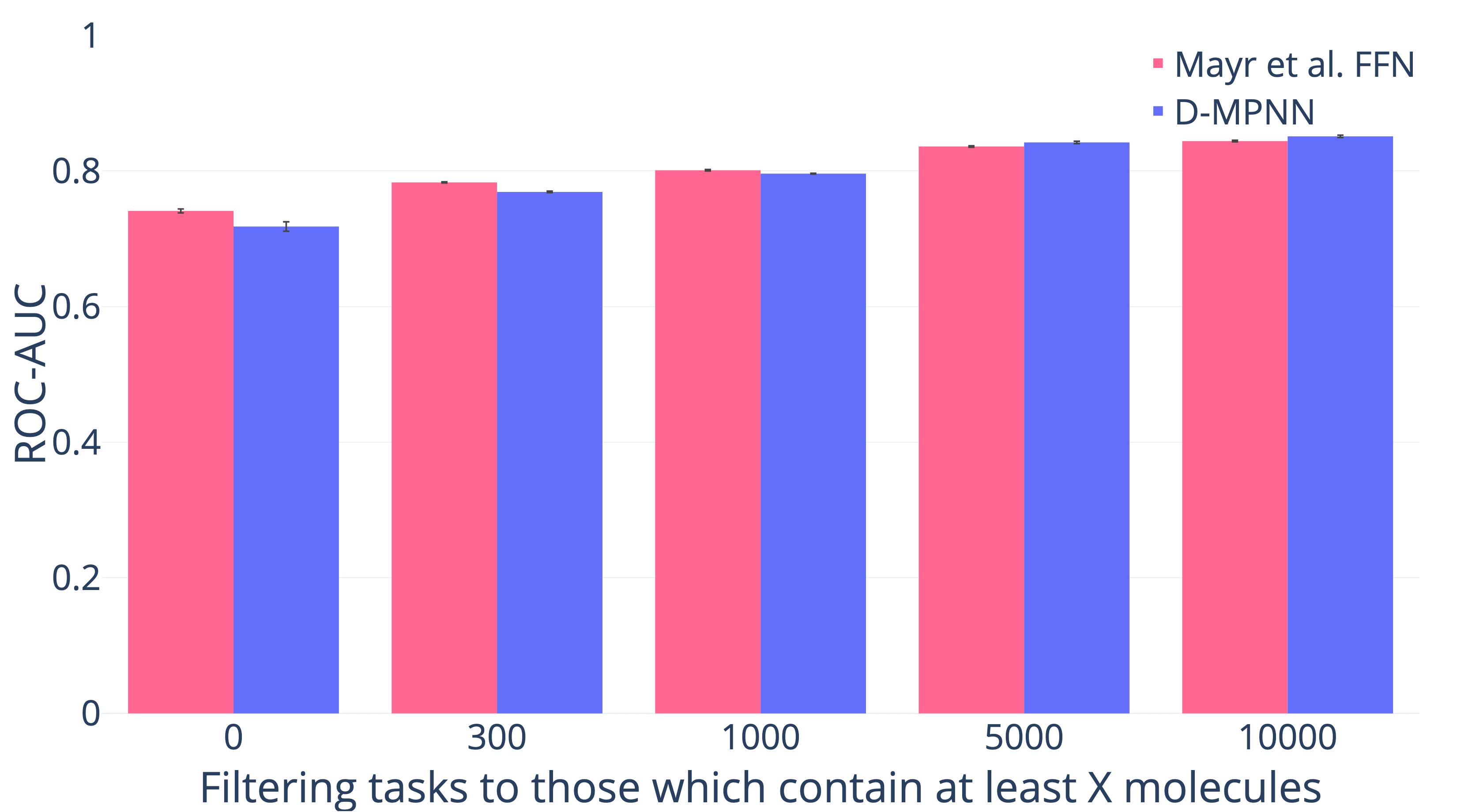}
    \caption{Effect of data size on the performance of the model from \citeauthor{mayr2018chembl}\cite{mayr2018chembl} and of our D-MPNN model (higher = better). All comparisons besides the first are statistically significant.}
    \label{fig:dataset_size}
\end{figure}

Finally, we analyze the effect of data size on the performance of our model, using the ChEMBL dataset. ChEMBL is a large dataset of 456,331 molecules on 1,310 targets, but is extremely sparse: only half of the 1,310 targets have at least $300$ labels. For this analysis, we use the original scaffold-based split of \citeauthor{mayr2018chembl}\cite{mayr2018chembl}, containing 3 cross-validation folds. From Figure \ref{fig:dataset_size}, we hypothesize that our D-MPNN struggles on low-label targets in comparison to this baseline. As our D-MPNN model does not use any human-engineered fingerprints or descriptors and must therefore learn its features completely from scratch based on the input data, it would be unsurprising if the average ROC-AUC score of D-MPNN is worse than that of the feed-forward network running on human-engineered descriptors in \citeauthor{mayr2018chembl}\cite{mayr2018chembl}. When we filter the ChEMBL dataset by pruning low-data targets at different thresholds, we find that our D-MPNN indeed may outperform the best model of \citeauthor{mayr2018chembl}\cite{mayr2018chembl} at larger data thresholds (Figure \ref{fig:dataset_size}), though our results are not fully conclusive.

\FloatBarrier

\section{Conclusion and Future Work}

\begin{table}[]
    \resizebox{\textwidth}{!}{
    \centering
    \begin{tabular}{|c|c|c|c|c|}
        \hline
        \textbf{Baseline} & \textbf{D-MPNN is \textit{better}} & \textbf{D-MPNN is \textit{same}} & \textbf{D-MPNN is \textit{worse}} & \textbf{\# datasets} \\
        \Xhline{2\arrayrulewidth}
        MoleculeNet\cite{Wu_2018} & 5 & 3 & 2 & 10 \\
        \citeauthor{mayr2018chembl}\cite{mayr2018chembl} & 8 & 10 & 1 & 19 \\
        RF on Morgan & 9 & 1 & 4 & 15 \\
        FFN on Morgan & 14 & 5 & 0 & 19 \\
        FFN on Morgan Counts & 15 & 4 & 0 & 19 \\
        FFN on RDKit & 8 & 5 & 4 & 19 \\
        \hline
    \end{tabular}}
    \caption{Number of public datasets where D-MPNN is statistically significantly better than, equivalent to, or worse than each baseline model.}
    \label{tab:significant_improvements}
\end{table}

In this paper, we performed an extensive comparison of molecular property prediction models based on either fixed descriptors or learned molecular representations by performing over 850 experiments on 19 public and 16 proprietary datasets. Table \ref{tab:significant_improvements} shows a summary of how our D-MPNN compares to each of the baseline models. Our model consistently matches or outperforms each baseline individually, and across all baselines, our model achieves comparable or better performance on 11 of the 19 public datasets: QM7, QM8, QM9, ESOL, FreeSolv, Lipophilicity, BBBP, PDBbind-F, PCBA, Tox21, and ClinTox. On the remaining 8 datasets, no single baseline model is consistently superior. Furthermore, our model's strong results transfer to proprietary datasets, where our model outperforms the random forest, feed-forward neural network, and \citeauthor{mayr2018chembl}\cite{mayr2018chembl} models on 15 out of the 16 datasets. The strong performance of our model over these baselines, many of which use computed fingerprints or descriptors, demonstrate that learned molecular representations are indeed ready for ``prime time'' use in industrial property prediction settings.

Nevertheless, several avenues for future research remain. When analyzing the performance of our D-MPNN, we found that it typically underperforms when either 1) the other models incorporate 3D information, as in MoleculeNet's best QM and PDBbind models, 2) the dataset is especially small, as in the PDBbind-C dataset, or 3) the classes are particularly imbalanced, as in the MUV dataset. One avenue of improvement is the incorporation of additional 3D information into our model, which currently includes only a very restricted and naive representation of such features. Another potential improvement is a principled pretraining approach, which some authors have already begun to explore\cite{navarin2018pre,goh2018using}. Such an approach could enable models to transfer learning from large chemical datasets to much smaller datasets, thereby improving performance in limited data settings. Another direction for future research is to determine how to adapt models and training algorithms to classification datasets with extreme class imbalance. Finally, in addition to these potential improvements, our analysis of how estimation of model generalizability is affected by split type opens the door to future work in uncertainty quantification and domain of applicability assessment. 

% Finally, we plan to add confidence estimation of our model's predictions on regression tasks in the near future, in order to increase our model's utility to chemists. %We leave these improvements to future work.
% CWC: Don't need to say "we leave these improvements to future work" so much
% New CWC: I think "we plan to add..." could invite criticism. Perhaps something more like "This increased understanding of how estimation of model generalizability is affected by scaffold split opens door to future work in uncertainty quantification and domain of applicability assessment"

\begin{acknowledgement}

We thank the Machine Learning for Pharmaceutical Discovery and Synthesis (MLPDS) consortium, Amgen Inc., BASF, and Novartis for funding this research. The MIT authors are funded by MIT, in particular the MLPDS consortium; the authors at companies are funded by their respective organizations. This work was supported by the DARPA Make-It program under contract ARO W911NF-16-2-0023. The authors declare no competing financial interest.

We would like to thank the other members of the computer science and chemical engineering groups in the Machine Learning for Pharmaceutical Discovery and Synthesis consortium for their helpful feedback throughout the research process. We would also like to thank the other industry members of the consortium for useful discussions regarding how to use our model in a real-world setting. We thank Nadine Schneider and Niko Fechner at Novartis for helping to analyze our model on internal Novartis data and for feedback on the manuscript. We thank Ryan White, Stephanie Geuns-Meyer, and Florian Boulnois at Amgen Inc. for their help enabling us to run experiments on Amgen datasets. In addition, we thank Lior Hirschfeld for his work on our web-based user interface\cite{chemprop_website}. Finally, we thank Zhenqin Wu for his aid in recreating the original data splits of \citeauthor{Wu_2018}\cite{Wu_2018}, and we thank Andreas Mayr for helpful suggestions regarding adapting the model from \citeauthor{mayr2018chembl}\cite{mayr2018chembl} to new classification and regression datasets. 

\end{acknowledgement}

\begin{suppinfo}

In our Supporting Information, we provide links to our code and to a demonstration of our web-based user interface. We also provide further comparisons of model performance on both scaffold-based and random splits of the data, and we provide tables with all raw performance numbers (including p-values) which appear in the charts in this paper. In addition, we analyze the class balance of classification datasets, and we provide a list of the RDKit calculated features used by our model.

\end{suppinfo}


\bibliography{paper}

%% graphical table of contents.
%%%%%%%%%%%%%%%%%%%%%%%%%%%%%%%%%%%%%%%%%%%%%%%%%%%%%%%%%%%%%%%%%%%%%

\begin{tocentry}

\centering
\includegraphics[width=1.0\linewidth]{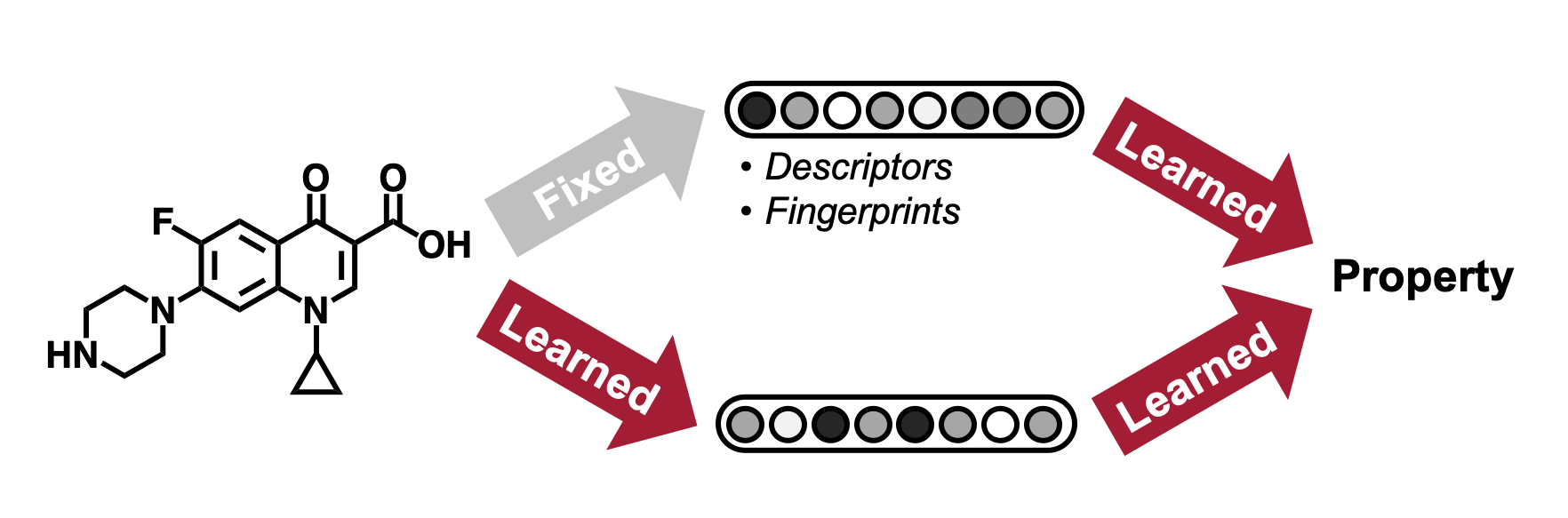}

\end{tocentry}

\end{document}